\documentclass[10pt,twocolumn,letterpaper]{article}

\usepackage{iccv}              

\definecolor{iccvblue}{rgb}{0.21,0.49,0.74}
\usepackage[pagebackref,breaklinks,colorlinks,allcolors=iccvblue]{hyperref}

\usepackage{tikz}
\usepackage{array}
\usepackage{times}
\usepackage{helvet}
\usepackage{amssymb}
\usepackage{amsthm}
\usepackage{amsmath}
\usepackage{amssymb}
\usepackage{courier}
\usepackage{multirow}
\usepackage{mathrsfs}
\usepackage{graphicx}
\usepackage{enumitem}
\usepackage{graphicx}
\usepackage{blindtext}
\usepackage{algorithm}
\usepackage{algorithmic}
\usepackage{lineno,hyperref}
\usepackage[marginal]{footmisc}
\usepackage{amsthm}
\usepackage[utf8]{inputenc}
\usepackage[english]{babel}
\usepackage{epstopdf}
\usepackage{array}
\usepackage{multirow}
\usepackage{booktabs}
\usepackage{dsfont}
\usepackage{comment}
\usepackage{endnotes}
\usepackage{makecell}
\usepackage{listings}
\usepackage{xcolor}
\usepackage{bm}

\def\paperID{4766} 
\def\confName{ICCV}
\def\confYear{2025}

\title{Generalized Deep Multi-view Clustering via Causal Learning with Partially Aligned Cross-view Correspondence}

\author{
Xihong Yang$^{1,3}$, Siwei Wang$^{2}$, Jiaqi Jin$^1$, Fangdi Wang$^1$, Tianrui Liu$^{1}$,\\ , Yueming Jin$^{3}$, Xinwang Liu$^{1,}$\thanks{Corresponding author}\space, En Zhu$^{1,*}$, Kunlun He$^4$
\\
$^1$College of Computer Science and Technology, National University of Defense Technology, Changsha, \\China
$^2$Intelligent Game and Decision Lab, Beijing, China\\
$^3$National University of Singapore, Singapore \quad 
$^4$Chinese PLA General Hospital, Beijing, China
\\
{\tt\small \{yangxihong, xinwangliu, enzhu\}@nudt.edu.cn}\\
}

\begin{document}
\maketitle

\begin{abstract}
Multi-view clustering (MVC) aims to explore the common clustering structure across multiple views. Many existing MVC methods heavily rely on the assumption of view consistency, where alignments for corresponding samples across different views are ordered in advance. However, real-world scenarios often present a challenge as only partial data is consistently aligned across different views, restricting the overall clustering performance. In this work, we consider the model performance decreasing phenomenon caused by data order shift (i.e., from fully to partially aligned) as a generalized multi-view clustering problem. To tackle this problem, we design a \textbf{cau}sal \textbf{m}ulti-\textbf{v}iew \textbf{c}lustering network, termed \textbf{CauMVC}. We adopt a causal modeling approach to understand multi-view clustering procedure. To be specific, we formulate the partially aligned data as an intervention and multi-view clustering with partially aligned data as an post-intervention inference. However, obtaining invariant features directly can be challenging. Thus, we design a Variational Auto-Encoder for causal learning by incorporating an encoder from existing information to estimate the invariant features. Moreover, a decoder is designed to perform the post-intervention inference. Lastly, we design a contrastive regularizer to capture sample correlations. To the best of our knowledge, this paper is the first work to deal generalized multi-view clustering via causal learning. Empirical experiments on both fully and partially aligned data illustrate the strong generalization and effectiveness of CauMVC.  
\end{abstract}

\section{Introduction}
Multi-view clustering (MVC)~\cite{sun2024robust,MFLVC,dong2,wan1,wan2,wan3,gao1,gao2,jin1,jin2,cai2021learning,DealMVC,zheng2024asymmetric,AIRMVC,yu1,yu2,yu3} is a critical task for uncovering semantic information and partitioning data into distinct groups in an unsupervised manner, gaining significant attention in recent years.

\begin{figure}[t]
\centering
\includegraphics[scale=0.32]{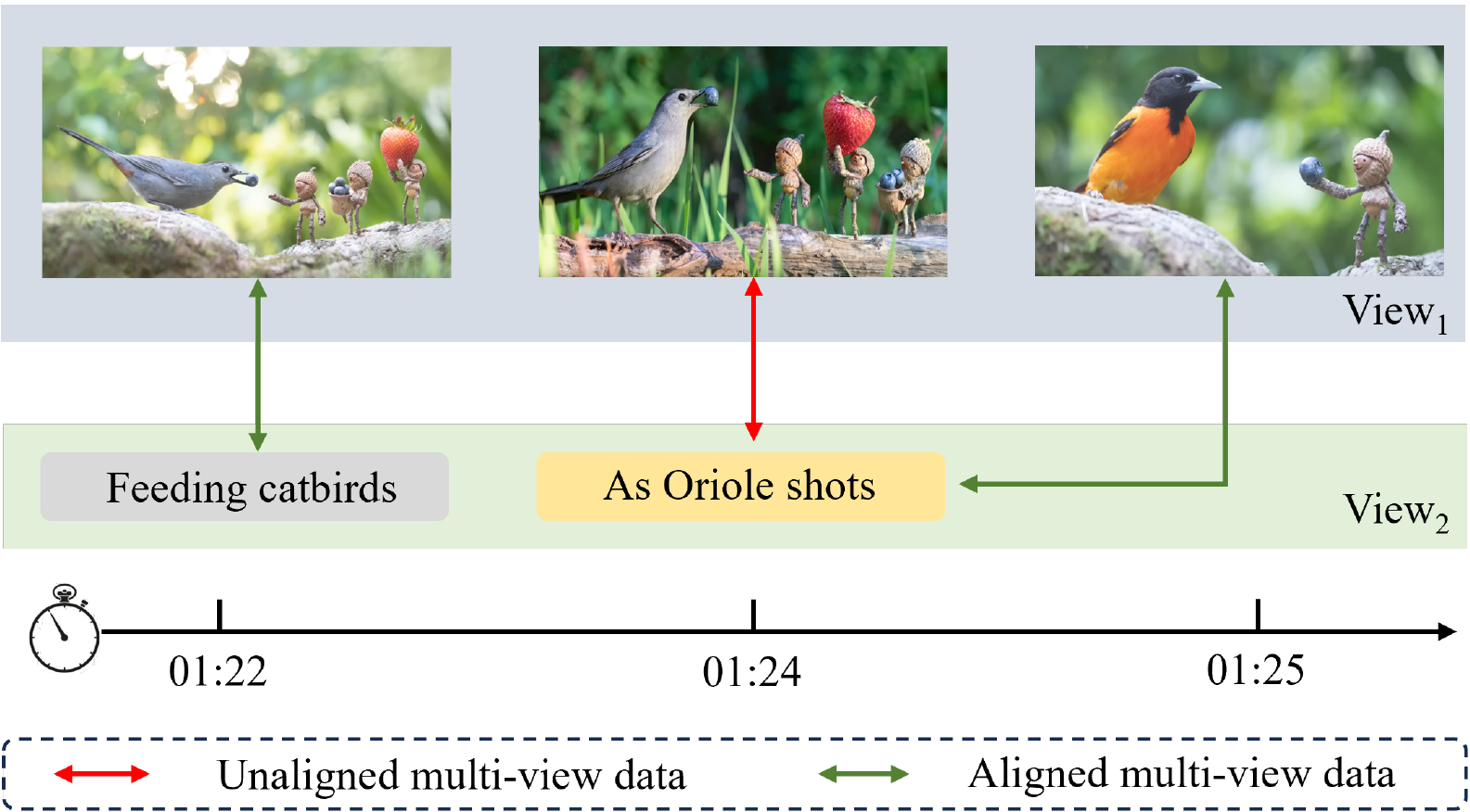}
\caption{An illustrative example of the motivation is presented through a raw video and its corresponding description. In this example, the issue of partial alignment between the video and the description becomes evident. In this scenario, we consider the video as one view and the description as another. There is a noticeable mismatch between the description and the visual content. For instance, when the description mentions “Oriole,” the video content actually shows “feeding catbirds” at the timestamp “01:24”.}
\label{motivation}  
\end{figure}

Although existing MVC algorithms demonstrate promising clustering performance, they heavily rely on the assumption that the mapping of all corresponding samples between views is ordered. This reliance facilitates learning a common representation and makes clustering feasible. However, in real application, the multi-view data is always partially aligned. To be concrete, taking a multi-view dataset $\{\textbf{X}^1, \textbf{X}^2\}$ as an example, the data $x_i^1$ and $x_i^2$ with the same index in two views may represent different content. The generalization and performance of MVC models diminish when erroneous alignment information between views misleads the final clustering outcome~\cite{MVC-UM}. As illustrated in Fig.~\ref{motivation}, the video content and its corresponding text description represent two distinct views. In this scenario, the text description does not synchronize with the video content\footnote{Video is collected from https://www.youtube.com/watch?v=BwXWg3VipQE.}. Recently, a serious multi-view clustering are proposed to solve this problem. MVC-UM~\cite{MVC-UM} employed non-negative matrix factorization to address unknown cross-view mappings. Besides, PVC~\cite{PVC} utilized a differentiable surrogate for the Hungarian algorithm to establish class-level correspondences in unaligned data, while SURE~\cite{SURE} introduced a noise-robust contrastive loss function. Despite achieving promising clustering performance, these algorithms overlook the causality between fully aligned and partially aligned data, lacking a unified causal framework to address the partially aligned problem

\begin{figure}[t]
\centering
\includegraphics[scale=0.32]{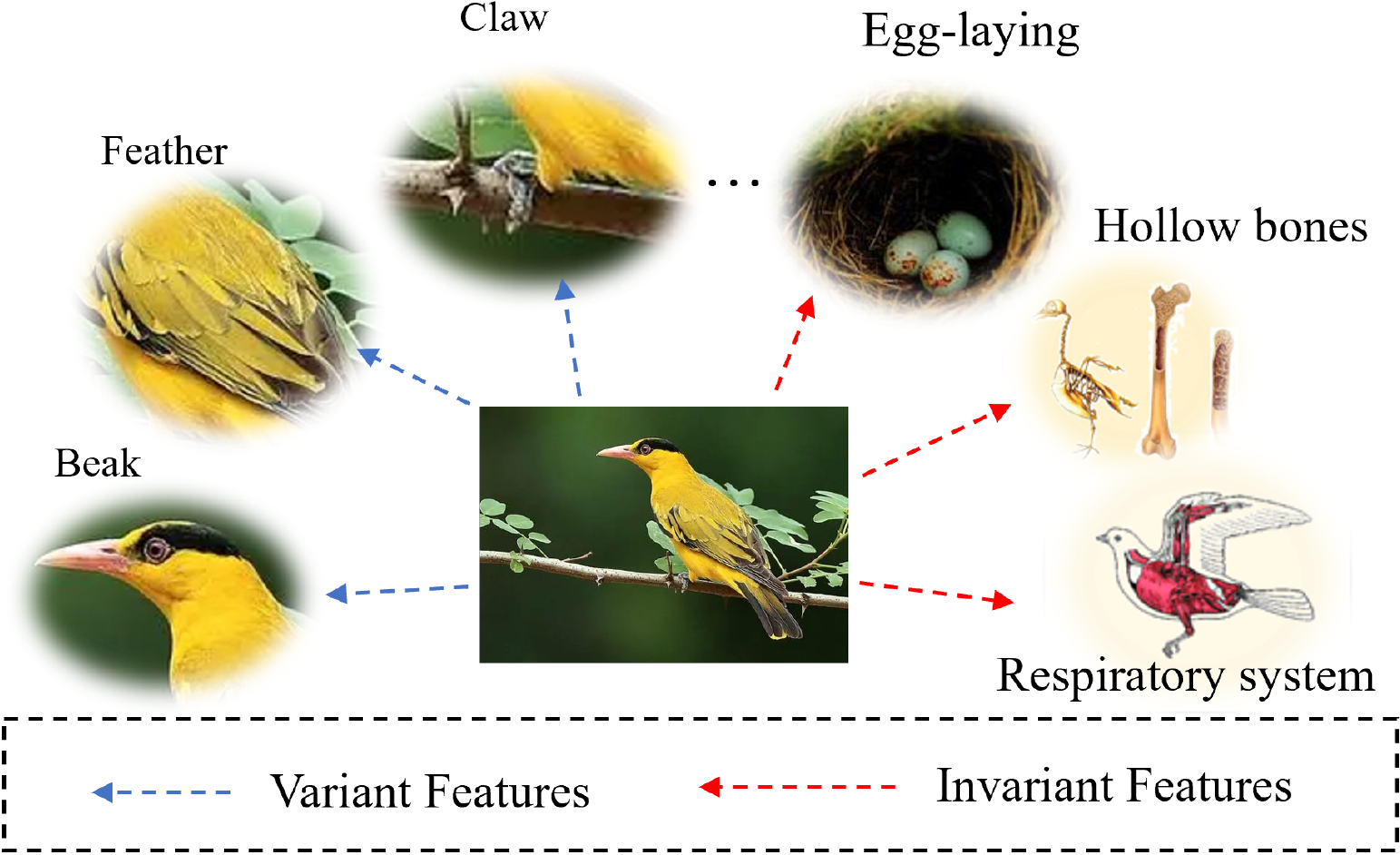}
\caption{Illustration of the variant features and invariant features. Taking the bird for an example, the beak, feathers, and claws are considered variant features. In contrast, characteristics such as egg-laying, hollow bones, and other biosignatures are considered invariant features, which remain unchanged across different samples in the dataset.}
\label{va_in_fea}  
\end{figure}

In this work, we consider the partially aligned data as the data shift. The objective is to realize strong generalization of the model. To achieve the above objective, we leverage causal method to delve into the cause-effect dynamics within the multi-view clustering process. We categorize the features into variant features $x_{va}$ and invariant features $x_{in}$. Here, we consider the input raw features as the variant features since those features are different for cross-view scenario. The invariant features conflict the fundamental attributes of the data, which keep consistent in different views. Taking the example in Fig.~\ref{va_in_fea} for illustrarion, for example, we demonstrate the variant and invariant features of a bird. Further, we classify the representations into two classes based on whether they are influenced by the variant features or not. The causal graph depicted in Fig.~\ref{causal_graph} illustrates the relationships from variant features $x_{va}$ and invariant features $x_{in}$ to the extracted variant representations $e_{va}$ and invariant representations $e_{in}$, culminating in the clustering result $r$. Existing methods extract representations by encoding the original multi-view raw feature inputs, i.e., variant feature $x_{va}$. When the input features drift from fully aligned data $x_{va}$ to partially aligned data $x'_{va}$, this paradigm will fail~\cite{SURE,PVC}. From the causal perspective, MVC with partially aligned data is viewed as the post-intervention inference result probabilities $p(r|do(x_{va}=x'_{va}), x_{in})$, where the input shift from $x_{va}$ to $x'_{va}$ is formulated as intervention. 

To be specific, we design a \underline{\textbf{Cau}}sal \underline{\textbf{M}}ulti-\underline{\textbf{V}}iew \underline{\textbf{C}}lustering network to realize the strong generalization with partially aligned data, termed CauMVC. The challenge is to obtain the invariant features $x_{in}$ essential for estimating $p(r|do(x_{va}=x'_{va}), x_{in})$. To overcome this problem, we employ a variational autoencoder (VAE) for inferring the invariant features $x_{in}$ through variational inference. Subsequently, a decoder network is introduced to estimate the clustering results, enabling post-intervention inference with partially aligned data features $x'_{va}$. Moreover, we incorporate a contrastive regularizer to explore correlations between samples, thereby enhancing the model's discriminative capabilities. Comprehensive experiments on both fully and partially aligned datasets validate the efficacy and generalizability of our CauMVC model.

The primary contributions of our proposed CauMVC can be summarized as follows.

\begin{itemize}
\item We consider the model performance decreasing caused by data shift (i.e., from fully to partially aligned) as a generalized problem, formulating and solving it from causal perspective.

\item We design a causal multi-view clustering network, employing causal modeling and inference to tackle the shift in multi-view input. 

\item Extensive experimentation on both fully and partially aligned data across eight benchmark datasets show the effectiveness and generalization of our method. 

\end{itemize}


\begin{figure*}
\centering
\scalebox{0.6}{
\includegraphics{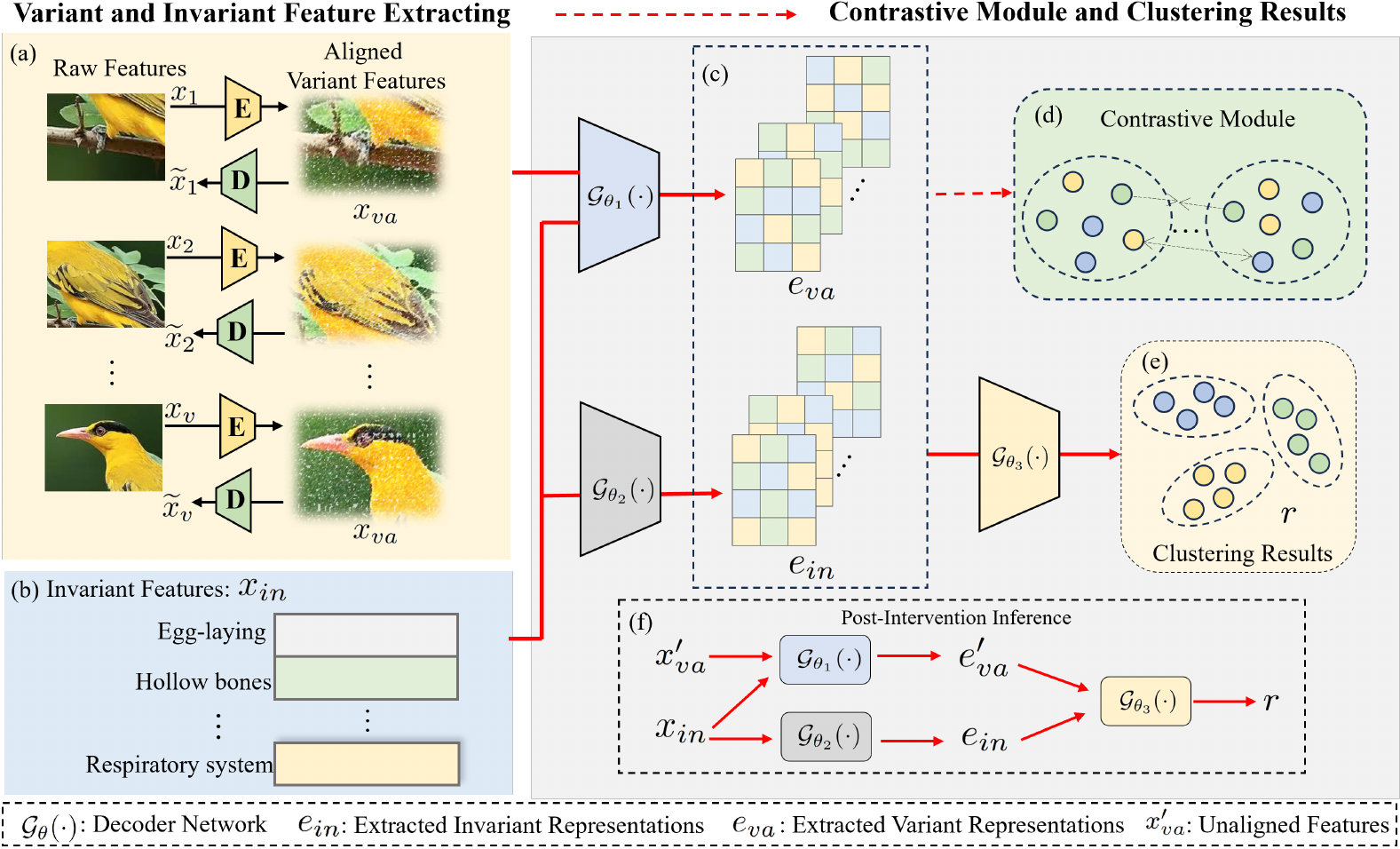}}
\caption{Illustration of the proposed causal multi-view clustering network. To be specific, we first design a autoencoder module (a) to obtain the variant features $x_{va}$ and reconstruct the input raw features. Besides, the invariant features $x_{in}$ are generated by the variantional auto-encoder with the variant features and clustering results. Then, we construct the causal graph based on the variant features $x_{va}$ and invariant features $x_{in}$. The variant and invariant representations are extracted by the network $\mathcal{G}_{\theta_1}(\cdot)$ and $\mathcal{G}_{\theta_2}(\cdot)$, respectively. Moreover, we design a contrastive module (d) to improve the discriminative capacity of the causal model. For the unaligned data, we utilize the causal graph to process the unaligned variant features $x'_{va}$ and the invariant features $x_{in}$ to generate the final clustering results $r$ with performing the post-intervention inference.} 
\label{overall}
\end{figure*}

\section{Preliminary}
In this paper, we address the multi-view clustering task in scenarios with partial alignment. The raw features from different views represent various descriptions of the same sample. Therefore, in this setting, we consider the aligned data as the variant feature for the same sample cross different views. For a given dataset $\{x^{(v)}\}_{v=1}^{V}$, we randomly spit the dataset into two partitions with 50\% ratio, i.e., the aligned data and unaligned data. Let $x_{va}$ and $x_{in}$ denote the variant and invariant features, respectively. Here, we consider the unaligned data as the shift phenomenon compared with the aligned scenario, which is denoted as $x'_{va}$. The invariant features $x_{in}$ are obtained by the encoder network $\mathcal{F}_{\phi(\cdot)}$. Then, we obtain the extracted variant representations $e_{va}$ and invariant representations $e_{in}$ by the encoder networks $\mathcal{G}_{\theta_1}$ and $\mathcal{G}_{\theta_2}$. Moreover, we define the clustering results as $r$. The fundamental notations used in this paper are outlined in Tab.~\ref{notation_table}.

\section{Related Work}
\label{Related Work} 

\subsection{Multi-view Clustering}

Recently, Multi-view Clustering (MVC) has garnered significant attention \cite{suyuan_TKDE,huangdong1, DealMVC, renyazhou,wan1,zheng2024asymmetric,AIRMVC}. Existing MVC methods can be broadly categorized into two groups based on cross-view correspondence: MVC with fully aligned data and MVC with partially aligned data. Fully aligned data implies predefined mapping relationships for every pair of cross-view data. There are several works under this assumption, which can be encompassed in five main categories: (1) Non-negative matrix factorization-based MVC \cite{wenjie_mf} aims to identify a shared latent factor, which is used to process information from multi-view input. (2) Kernel learning-based MVC \cite{one, simplemkkm} involves predefining a base kernels set for each views. After that, this method optimally fuse the weights of the kernels to improve clustering outcomes. (3) Subspace-based MVC \cite{suyuan_AAAI} is based on the assumption that all views in the multi-view task share a low-dimensional latent space, with the final outcomes derived from learning this shared representation. (4) Benefiting from the successful of graph learning and its applications~\cite{MGCN,CCGC,xihong,Darec,CONVERT,DCRN,Graphlearner,HSAN,Dink_net,yin2024dataset,TTTRec}, Graph-based MVC \cite{graph_mvc} seeks to constructing a unified graph from multiple views, with clustering results derived from spectral decomposition. (5) Thanks to the robust representational capabilities of deep networks~\cite{zhangyang1,zhangyang2, zhangyang3,niu1,niu2}, deep neural network-based MVC \cite{DealMVC,jiaqi_cvpr,jinjiaqi2,dai} has the capacity to extract more sophisticated representations. through neural networks. Despite achieving promising clustering performance, most of these methods heavily rely on the assumption that cross-view data are fully aligned.

\begin{table}[]
\centering
\scalebox{1.0}{
\begin{tabular}{cc}
\hline
Notation                                                                      & Meaning                             \\ \hline
$x_{va} $                                                                    & Fully Aligned Variant Features      \\
$x_{in}$                                                                     & Invariant Features                  \\
$x'_{va}$                                                                    & Partially Aligned Variant Features  \\
$e_{va}$                                                                     & Extracted Variant Representations   \\
$e_{in} $                                                                    & Extracted Invariant Representations \\
$r   $                                                                          & Clustering Results                  \\
$\mathcal{F}_{\phi}(\cdot)$   & Variantional Auto-Encoder Network   \\
$\mathcal{G}_{\theta}(\cdot)$ & Post-Intervention Inference Network \\
$V$                                                                             & The Number of Views                 \\
$N $                                                                            & The Number of Samples               \\
$Z$                                                                             & The Similarity Matrix               \\ \hline
\end{tabular}}
\caption{Basic notations used in the whole paper.}
\label{notation_table}
\end{table}

To tackle this issue, many MVC algorithms have been proposed \cite{PVC, SURE, MvCLN,wenyi}. PVC is designed to use a differentiable surrogate of the non-differentiable Hungarian algorithm to learn the correspondence of partially aligned data. MVC-UM \cite{MVC-UM}, based on non-negative matrix factorization, learns the correspondence by exploring cross-view relationships. SURE \cite{SURE} uses available pairs as positives and randomly selects some cross-view samples as negatives. UPMGC-SM \cite{wenyi} leverages structural information from each view to refine cross-view correspondences. In contrast to the above methods, we approach partially aligned data from a causal perspective, aiming to improve the generalization of the model.

\subsection{Causal Disentangled Representation Learning}
Traditional approaches for disentangled representation learning focus on examining mutually independent latent factors through the use of encoder-decoder networks. In this approach, a standard normal distribution is utilized as the prior for the latent code. Moreover, the variational posterior $q(z|x)$ is employed to approximate the unknown posterior $p(z|x)$. $\beta$-VAE \cite{beta_vae} introduces an adaptive framework to adjust the weight of the \text{KL} term. Factor VAE \cite{factor_vae} designs a framework, which focuses solely on the independence of factors. After that, the exploration of causal graphs from observations has gained significant attention, leveraging either purely observational data or a combination of observational and interventional data. NOTEARs \cite{NOTEARs} incorporates a novel Directed Acyclic Graph (DAG) constraint for causal learning. LiNGAM \cite{LiNGAM} ensures the identifiability of the model based on the assumptions of linear relationships and non-Gaussianity. In cases where interventions are feasible, Heckerman et al. \cite{byes} demonstrate the causal structure learned from interventional data can be identified. More recently, there has been an increasing interest in combining causality and disentangled representation. Suter et al. \cite{Suter} employs causality to explain disentangled latent representations, while Kocaoglu et al. \cite{causalgan} introduces CausalGAN, a method supporting "do-operations" on images. Drawing inspiration from the success of causal learning, we apply causal modeling to multi-view clustering. To the best of our knowledge, our work represents the first attempt to leverage causal learning to improve model generalization with partially aligned data in the multi-view clustering task.

\section{Method}

In this section, we introduce a generalized multi-view clustering algorithm from a causal perspective. Specifically, we first present the relevant notations and formulations. Subsequently, We provide a detailed explanation of the causal multi-view clustering model and the contrastive regularizer. Finally, we introduce the training loss function of proposed CauMVC. The framework of CauMVC is shown in Fig.~\ref{overall}.

\subsection{Problem Definition}
 
\subsubsection{Multi-view Clustering from Causal Perspective}

In this paper, the objective is to achieve strong generalization in a partially aligned data environment via causal learning. We abstract the multi-view clustering process as a causal graph, shown in Fig.\ref{causal_graph}. We provide the preliminary and symbol definitions in the Appendix. In following, we explain the rationality. 
\begin{itemize}
    \item $x_{va}, x_{in}$ represent the variant features and invariant features, respectively. For instance, in the context of a bird dataset, the variant features include characteristics like the beak, feathers, and claws. invariant features can be considered as consistent features across different views, such as egg-laying animals, hollow bones, and respiratory system.
    \item $e_{va}$ and $e_{in}$ denote the extracted variant and invariant representations. $e_{in}$ is isolated because there always exist representations that remain unaffected by $x_{va}$. For instance, the hollowness of bird bones is unrelated to the presence of feathers.
    \item ${r}$ is the multi-view clustering results.
    \item $(x_{va}, x_{in}) \longrightarrow e_{va}$ and $x_{in} \longrightarrow e_{in}$ mean the extracted representations are determined by the input features.
    \item $(e_{va}, e_{in}) \longrightarrow {r}$ is that the multi-view clustering task is affected by the extracted representations. 
\end{itemize}

\subsubsection{Multi-view Clustering with Partially Aligned Data}
Multi-view clustering (MVC) seeks to uncover the shared clustering structure among various views, segmenting the multi-view data into distinct clusters in an unsupervised manner. In this work, we explore multi-view clustering task, where the input multi-view data is shift from aligned data $x_{va}$ to partially aligned data $x'_{va}$. From the causal view, we termed the data shift as an intervention~\cite{intervention}, which can be presented as $do(x_{va} = x'_{va})$. The primary objective is for the model to generalize and produce the post-intervention distribution of clustering.

\subsection{Causal Multi-view Clustering Module}

\begin{figure}[t]
\centering
\includegraphics[scale=0.22]{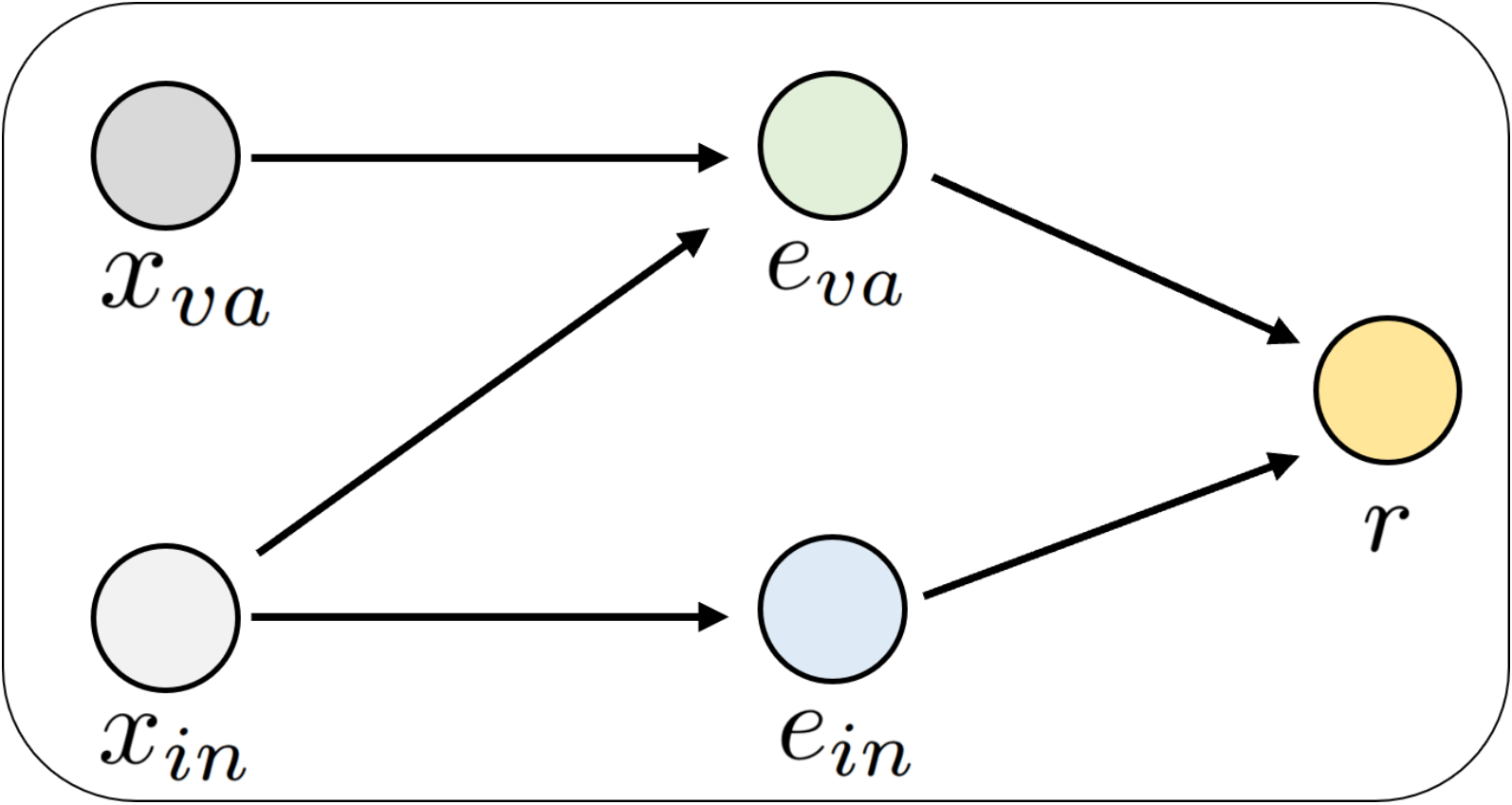}
\caption{Causal graph illustrating the process of multi-view clustering. To be specific, $x_{va}$ and $x_{in}$ denote the variant features and invariant features, respectively. Then, the extracted variant and invariant representations $e_{va}$ and $e_{in}$ are obtained by the encoder network. Finally, the clustering results $r$ are generated from $e_{va}$ and $e_{in}$.}
\label{causal_graph}  
\end{figure}

We consider the learning process illustrated in Fig.\ref{causal_graph} to construct the multi-view clustering model. To be specific, the model commences by sampling a d-dimensional invariant features $x_{in}$ from a standard Gaussian prior~\cite{guassian2,guassian}. From variant features $x_{va}$ and invariant features $x_{in}$, we embed features to latent space and use $e_{va}$ and $e_{in}$ to produce multi-view clustering. Inspired by the prior works~\cite{guassian, guassian1, guassian2}, we assume the features and the clustering results follow factorized Gaussian and multinomial priors, respectively. The distribution of $x_{in}, e_{va}, e_{in}$ and $r$ could be expressed as:  

\begin{equation} 
\begin{aligned}
x_{in} &\sim \mathcal{N}(0, \textbf{I}_d), \\
e_{va} &\sim \mathcal{N}(\mu_{\theta_1}(x_{va}, x_{in}), diag\{\sigma^2_{\theta_1}(x_{va}, x_{in})\}),\\
e_{in} &\sim \mathcal{N}(\mu_{\theta_2}(x_{in}), diag\{\sigma^2_{\theta_2}(x_{in})\}),\\
r &\sim \text{Mult}(\text{R}, \pi(\mathcal{G}_{\theta_3}(e_{va}, e_{in}))).
\end{aligned}
\label{init}
\end{equation}
where $\theta_1(x_{va}, x_{in})$ represents the mean of the Gaussian distribution estimated from $x_{va}$ and $x_{in}$ by the function $\mathcal{G}_{\theta_1}(x_{va}, x_{in})$ parameterized by $\theta_1$. The term $diag\{\sigma^2_{\theta_1}(x_{va}, x_{in})\}$ represents the diagonal covariance of the Gaussian distribution. Similarly, $\mathcal{G}_{\theta_2}(x_{in})$ is employ to compute the mean and diagonal covariance of $e_{in}$. Moreover, the clustering results ${r}$ is drown from the multinomial distribution, which can be presented with parameters of $\text{R} = \sum_{i=1}^N{r_i} $. Besides, $\pi(\mathcal{G}_{\theta_3}(e_{va}, e_{in}))$. $\pi(\cdot)$ is the softmax function used to normalize the output of $\mathcal{G}_{\theta_3}(e_{va}, e_{in})$.


To optimize the parameters $\theta_1, \theta_2, \theta_3$, we reconstruct ${r}$ based on the variant features $x_{va}$. Concretely, for a multi-view data with $x_{va}$, we maximize the log-likelihood $\text{log}~ p(r|x_{va})$, which can be expressed as:
\begin{equation} 
\begin{aligned}
\text{log}~p(r|x_{va}) &= \text{log} \int p(r, x_{in}|x_{va})d{x_{in}}\\ 
&= \text{log} \int p(r|x_{va}, x_{in})p(x_{in})d{x_{in}}.
\end{aligned}
\label{origin}
\end{equation}

However, due to the intractability of integrating over invariant features $x_{in}$ in Eq. \eqref{origin}, we address this challenge by drawing inspiration from variational inference~\cite{guassian,cor}. Specifically, we introduce a variational distribution $q(x_{in}|\cdot)$ to derive the evidence lower bound (ELBO) of Eq. \eqref{origin} as:
\begin{equation} 
\begin{aligned}
\text{log}~p(r|x_{va}) &= \text{log} \int p(r|x_{va}, x_{in})p(x_{in})\frac{q(x_{in}|\cdot)}{q(x_{in}|\cdot)}d{x_{in}}\\ 
&\ge \mathbb{E}_{q(x_{in}|\cdot)}\left [ \text{log} \frac{p(r|x_{va}, x_{in})p(x_{in})}{q(x_{in}|\cdot)} \right ] \\
&= \mathbb{E}_{q(x_{in}|\cdot)}\left [ \text{log} p(r|x_{va}, x_{in}) \right ]\\ 
&-\text{KL}(q(x_{in}|\cdot)||p(x_{in})),
\end{aligned}
\label{elbo}
\end{equation}
where the first term in Eq.\eqref{elbo} aims to reconstruct while the second term is to regularize the Kullback-Leibler divergence between $q(x_{in}|\cdot)$ and the prior $p(x_{in})$. The goal is to maximize the ELBO in Eq.\eqref{elbo}, thereby increasing the log-likelihood in Eq.\eqref{origin}. To achieve this objective, we design the variantional auto-encoder and decoder network to model $q(x_{in}|\cdot)$ and $p(r|x_{va}, x_{in})$, respectively.

\subsubsection{Variantional Auto-Encoder}
Invariant features reflect the fundamental attributes of the data, which remain consistent regardless of changes in perspective. \textbf{However, how to obtain the invariant features is a challenging problem.} In this work, we attempt to infer the invariant features from the existing information. To be specific, we assume that the clustering results and the variant features are likely to indicate the invariant features. Taking the bird dataset for example, animals characterized by features such as feathers and beaks, and subsequently classified as birds through clustering algorithms, are likely to exhibit biological traits such as egg-laying and possessing hollow bones. Therefore, we define $q(x_{in}|\cdot) = q(x_{in}|r',x_{va})$. To efficiently estimate of $ q(x_{in}|r',x_{va})$, we employ amortized inference~\cite{Amortized_inference} and integrate an encoder network. This process can be described as follows:

\begin{equation} 
\begin{aligned}
q(x_{in}|r',x_{va}) = \mathcal{N}\left ( x_{in}; \mu_\varphi(r', x_{va}), \text{diag}\{ \sigma^2_\varphi(r', x_{va})\}  \right ), 
\end{aligned}
\label{encoder}
\end{equation}
where $\mu_\varphi(\cdot)$ and $\sigma_\varphi(\cdot)$ are the parameters obtained by the encoder network $\mathcal{F}_\varphi(\cdot)$, i.e., $\mathcal{F}_\varphi(r, x_{va}) = [\mu_\varphi(r, x_{va}), \sigma_\varphi(r, x_{va})]$. $r'$ is the clustering results obtained by the pretrain model. We employ the multi-layer perception (MLP) as encoder network $\mathcal{F}_\varphi(\cdot)$.

\subsubsection{Decoder Network}
We decompose $p(r|x_{va}, x_{in})$ according to Eq.\eqref{init}, which can be expressed as:
\begin{equation} 
\begin{aligned}
p(r|x_{va},x_{in}) = \int \int p(e_{va}|x_{va}, x_{in}) p(e_{in}|x_{in})p(r|e_{va},e_{in})d{e_{va}}d{e_{in}},
\end{aligned}
\label{decoder}
\end{equation}
where the parameters of $p(e_{va}|x_{va}, x_{in})$ and $p(e_{in}|x_{in})$ are determined by $\mathcal{G}_{\theta_1}(x_{va}, x_{in})$ and $\mathcal{G}_{\theta_2}(x_{in})$, respectively. Similar to encoder network, we employ decoder networks by two MLPs, denoted as $\mathcal{G}_{\theta_1}(x_{va},x_{in}) = [\mu_{\theta_1}(x_{va}, x_{in}), \sigma_{\theta_1}(x_{va}, x_{in})]$ and $\mathcal{G}_{\theta_2}= [\mu_{\theta_2}(x_{in})], \sigma_{\theta_2}(x_{in})$.

\subsubsection{Approximation Process}
Though the above analysis, we could obtain the estimations of $p(e_{va}|x_{va}, x_{in})$ and $p(e_{in}|x_{in})$. However, the calculation of $p(r|x_{va},x_{in})$ is challenging due to the computationally expensive integration over the variables $e_{va}$ and $e_{in}$. Consequently, we utilize Monte Carlo sampling~\cite{mc_sample} to enhance efficiency. Formally,
\begin{equation} 
\begin{aligned}
p(r|x_{va},x_{in}) \approx \frac{1}{A}\frac{1}{B} \sum_{a=1}^A \sum_{b=1}^B p(r|e_{va}^a, e_{in}^b),
\end{aligned}
\label{MC_sample}
\end{equation}
where $A$ and $B$ are the sample numbers. Besides, $e_{va}^a$ and $e_{in}^b$ are drawn from the distributions $p(e_{va}|x_{va}, x_{in})$ and $p(e_{in}|x_{in})$, respectively. However, the computationally cost associated with Eq.\eqref{MC_sample} remains high due to the necessity of calculating the conditional probability for $A \times B$ iterations. To address this challenge, we employ a widely adopted approximation technique~\cite{bounds,bounds1} as:
\begin{equation} 
\begin{aligned}
p(r|x_{va},x_{in}) \approx p \left ( r|\frac{1}{A}\sum_{a=1}^A e_{va}^a, \frac{1}{B}\sum_{b=1}^B e_{in}^b \right ) = p(r|\bar{e}_{va}, \bar{e}_{in}),
\end{aligned}
\label{MC_sample}
\end{equation}
where the approximation error (Jensen gap~\cite{Jensen_gap}) can be effectively bounded for most functions when calculating $p(r|\bar{e}_{va}, \bar{e}_{in})$~\cite{bounds1}. We employ an MLP as the backbone for $\mathcal{G}_{\theta_3}(\cdot)$. Therefore, we can estimate the parameters of $p(r|\bar{e}_{va}, \bar{e}_{in})$. The reconstruction term $\text{log}~p(r|x_{va}, x_{in})$ can be calculated as follows:
\begin{equation} 
\begin{aligned}
\text{log}~p(r|\bar{e}_{va}, \bar{e}_{in}) \overset{c}{=} \sum_{i=1}^N r'_i \text{log}~ \pi(\mathcal{G}_{\theta_3}(\bar{e}_{va}, \bar{e}_{in})),
\end{aligned}
\label{Final}
\end{equation}
where $r'_i$ represents the clustering results of the multi-view data obtained by the pretrain model. $\pi(\mathcal{G}_{\theta_3}(\cdot))$ signifies the prediction of the $i$-th sample after applying softmax normalization $\pi(\cdot)$ to the output of $\mathcal{G}_{\theta_3}(\cdot)$. Specifically, Eq.\eqref{Final} determines the reconstruction probability of obtaining $r$ from the multinomial distribution by performing $\text{R}$ times.


In summary, our goal is to maximize the Evidence Lower Bound (ELBO) in Eq.\eqref{elbo} in order to optimize $\varphi$ and $\theta = \{\theta_1, \theta_2, \theta_3\}$. Moreover, following~\cite{guassian, cor}, we utilize the KL annealing to regulate the KL divergence regularization. The ELBO loss function can be presented as:

\begin{equation} 
\begin{aligned}
\mathcal{L}_{ELBO} &= \mathbb{E}_{q_\varphi(x_{in}|r,x_{va})}[\text{log} p_{\theta}(r|x_{va},x_{in})]\\
&- \text{KL}(q_\varphi(x_{in}|r,x_{va})||p(x_{in})).
\end{aligned}
\label{loss_cau}
\end{equation}

\subsubsection{Causal Inference of Multi-view Clustering with Partially Aligned Data}
After that, we provide the explanation of how to infer the post-intervention probabilities $p(r|x'_{va}, x_{in})$, where $x'_{va}$ denotes the shift data, i.e., partially aligned multi-view data. 

Concretely, the causal inference is divided into two steps. Firstly, we acquire the invariant features based on Eq.\eqref{encoder}, which can be formulated as:
\begin{equation} 
\begin{aligned}
q(x_{in}|r,x'_{va}) = \mathcal{N}\left (x_{in}; \mu_\varphi(r, x'_{va}), \text{diag}\{ \sigma^2_\varphi(r, x'_{va})\}  \right ). 
\end{aligned}
\label{unobserved_feature}
\end{equation}

Then, according to Eq.\eqref{decoder}, we feed  $x'_{va}$ and $x_{in}$ to the decoder model to generate the post-intervention probabilities as:
\begin{equation}
\begin{aligned}
p(r|x'_{va},x_{in}) = \int \int p(e'_{va}|x'_{va}, x_{in}) p(e_{in}|x_{in})p(r|e'_{va},e_{in})d{e'_{va}}d{e_{in}},
\end{aligned}
\label{post_inference}
\end{equation}


\subsection{Contrastive Regularizer}

In this subsection, we design a contrastive regularizer to mine the relationship between samples. Specifically, we compute the similarity matrix $\textbf{Z}$ between the variant representations $e_{va}$ and invariant representations $e_{in}$:
\begin{equation}
\textbf{Z}= \frac{\left(e_{va}\right) \left(e_{in}\right)^{\mathsf{T}}}{{||e_{va}||}_2 {||e_{in} ||_2}}. 
\label{correlation}
\end{equation}

Then, we constrain the $\textbf{Z}$ to be close to an identity matrix $\textbf{I}$. The contrastive loss can be presented as:
\begin{equation}
\begin{aligned}
\mathcal{L}_C &= \frac{1}{N^2}\sum (\textbf{Z}-\textbf{I})^2 \\
 &=  \frac{1}{N}\sum\limits_{i=1}^N \left(\textbf{Z}_{ii}-1\right)^2
+
\frac{1}{N^2-N}\sum\limits_{i=1}^N \sum\limits_{j\ne i} \left(\textbf{Z}_{ij}\right)^2,   
\end{aligned}
\label{loss_con}
\end{equation}
where the initial term in Eq.\eqref{loss_con} compels the diagonal elements of $\textbf{Z}$ to 1. This operation encourages the representations of the same sample to agree with each other in latent space. Moreover, the latter term drives the off-diagonal elements to approach 0, thus pushing apart the representations of different samples.
\subsection{Loss Function}

Autoencoder\cite{dong1,DealMVC} is widely used in multi-view clustering. In this work, we utilize an autoencoder to learn the semantics across different views in the latent space. The reconstructed loss is calculated as:

\begin{equation} 
\mathcal{L}_R = \sum_{v=1}^V  \left \| x_{va}^v - \widetilde{x}_o^v \right \|_F^2,
\label{loss_r}
\end{equation}
where $x_{va}$ and $\widetilde{x}_o$ denote input observed multi-view feature and reconstructed features, respectively. The loss function of our proposed CauMVC mainly include three components, i.e., the reconstructed loss $\mathcal{L}_R$, the ELBO loss $\mathcal{L}_{ELBO}$ and the contrastive loss $\mathcal{L}_C$. Overall, the loss function can be calculated as:
\begin{equation} 
\mathcal{L} = \mathcal{L}_R + \alpha \mathcal{L}_{ELBO} + \beta \mathcal{L}_C,
\label{total_loss}
\end{equation}
where $\alpha$ and $\beta$ are the trade-off hyper-parameters. Detailed training procedure is presented in Alg.~\ref{ALGORITHM}.

\begin{algorithm}[t]
\small
\caption{Inference Pipeline of CauMVC with Partially Aligned Data}
\label{ALGORITHM}
\vspace{-10pt}
\flushleft{\textbf{Input}: The partially aligned data $x'_{va}$}; the interation number $I$
\vspace{-10pt}
\flushleft{\textbf{Output}: The clustering result ${r}$.} 
\begin{algorithmic}[1]
\vspace{-10pt}
\FOR{$i=1$ to $I$}
\STATE Obtain the invariant features $x_{in}$ by $\mathcal{F}_{\varphi(\cdot)}$ with Eq.~(10).
\STATE Encoder the representations $e'_{va}$ and $e_{in}$ by $\mathcal{G}_{\theta_1}$ and $\mathcal{G}_{\theta_2}$.
\STATE Obtain the post-intervention inference ${r}$ with Eq.~(11).
\STATE Calculate the ELBO loss, contrastive loss, and reconstruction loss with Eq.~(9), (13) and (14).
\STATE Calculate the total loss $\mathcal{L}$ by Eq.~(15).
\STATE Update model by minimizing $\mathcal{L}$ with Adam optimizer.
\ENDFOR
\STATE \textbf{return} ${r}$
\end{algorithmic}
\end{algorithm}


\begin{table*}[]
\centering
\scalebox{0.8}{
\begin{tabular}{c|cc|ccc|ccc|ccc|ccc}
\hline
\multirow{2}{*}{Aligned}   & \multicolumn{2}{c|}{Mthods}                                    & \multicolumn{3}{c|}{Movies}                                                        & \multicolumn{3}{c|}{UCI-digit}                                                     & \multicolumn{3}{c|}{STL10}                                                               & \multicolumn{3}{c}{SUNRGBD}                                                             \\ \cline{2-15} 
                           & \multicolumn{2}{c|}{Evaluation metrics}                        & ACC                       & NMI                       & PUR                        & ACC                       & NMI                       & PUR                        & ACC                       & NMI                             & PUR                        & ACC                       & NMI                       & PUR                             \\ \hline
\multirow{10}{*}{Fully}    & SDMVC                       & TKDE 2021                        & 27.71                     & 26.07                     & 30.28                      & 63.00                     & 64.29                     & 67.05                      & 30.01                     & 25.27                           & 31.07                      & 17.22                     & 14.11                     & 21.53                           \\
                           & CoMVC                       & CVPR2021                         & 17.50                     & 14.45                     & 19.45                      & 46.40                     & 49.36                     & 46.45                      & 23.55                     & 16.26                           & 25.01                      & 16.03                     & 11.59                     & 21.60                           \\
                           & SiMVC                       & CVPR2021                         & 17.18                     & 14.64                     & 18.80                      & 20.75                     & 16.59                     & 21.75                      & 16.04                     & 06.34                           & 17.02                      & 13.44                     & 10.52                     & 21.63                           \\
                           & SDSNE                       & AAAI 2022                        & {\underline{29.34}}               & {\underline{26.32}}               & {\underline{30.31}}                & 84.55                     & \textbf{89.05}            & {\underline{87.05}}                & O/M                       & O/M                             & O/M                        & O/M                       & O/M                       & O/M                             \\
                           & MFLVC                       & CVPR 2022                        & 20.75                     & 22.05                     & 23.99                      & {\underline{87.00}}               & 80.40                     & 87.00                      & 31.14                     & 25.36                           & 31.25                      & 13.22                     & 13.69                     & 23.12                           \\
                           & DSMVC                       & CVPR 2022                        & 17.02                     & 12.60                     & 19.12                      & 85.45                     & 80.67                     & 85.40                      & 27.53                     & 19.33                           & 29.41                      & 17.76                     & 10.43                     & 10.50                           \\
                           & GCFAggMVC                   & CVPR 2023                        & 26.42                     & 25.43                     & 26.98                      & 81.55                     & 77.70                     & 81.55                      & {\underline{36.69}}               & 28.78                           & {\underline{38.36}}                & 17.98                     & {\underline{13.86}}               & 22.64                           \\
                           & DealMVC                     & ACM MM 2023                      & 26.42                     & 23.79                     & 28.20                      & 86.20                     & 81.78                     & 86.20                      & 36.44                     & 29.93                           & 36.95                      & {\underline{18.16}}               & 08.58                     & 19.17                           \\
                           & \multicolumn{1}{l}{TGM-MVC} & \multicolumn{1}{l|}{ACM MM 2024} & \multicolumn{1}{l}{17.18} & \multicolumn{1}{l}{16.80} & \multicolumn{1}{l|}{18.96} & \multicolumn{1}{l}{54.35} & \multicolumn{1}{l}{62.76} & \multicolumn{1}{l|}{59.40} & \multicolumn{1}{l}{26.18} & \multicolumn{1}{l}{{\underline{30.86}}} & \multicolumn{1}{l|}{27.51} & \multicolumn{1}{l}{15.37} & \multicolumn{1}{l}{13.80} & \multicolumn{1}{l}{{\underline{23.20}}} \\
                           & CauMVC                      & Ours                             & \textbf{29.98}            & \textbf{26.41}            & \textbf{30.96}             & \textbf{91.85}            & {\underline{85.90}}               & \textbf{91.85}             & \textbf{37.88}            & \textbf{31.30}                  & \textbf{38.95}             & \textbf{18.83}            & \textbf{14.68}            & \textbf{24.20}                  \\ \hline
\multirow{6}{*}{Partially} & PVC                         & NeurIPS 2020                     & 14.26                     & 13.29                     & {\underline{16.53}}                & {\underline{77.30}}               & {\underline{75.54}}               & {\underline{80.50}}                & {\underline{26.18}}               & {\underline{25.22}}                     & {\underline{31.33}}                & 10.71                     & 11.28                     & {\underline{19.33}}                     \\
                           & MvGLN                       & CVPR 2021                        & 14.39                     & 10.87                     & 15.79                      & 66.78                     & 58.93                     & 67.48                      & 10.00                     & 09.06                           & 10.00                      & 09.97                     & 09.79                     & 11.40                           \\
                           & MVC-UM                      & KDD 2021                         & {\underline{23.76}}               & {\underline{26.42}}               & 13.50                      & 61.09                     & 56.61                     & 63.60                      & O/M                       & O/M                             & O/M                        & 10.34                     & 12.93                     & 18.73                           \\
                           & SURE                        & TPAMI 2023                       & 14.82                     & 11.92                     & 16.24                      & 51.76                     & 37.64                     & 52.56                      & 23.82                     & 13.20                           & 25.00                      & 11.11                     & 13.11                     & 16.43                           \\
                           & RMCNC                       & TKDE 2024                        & 14.10                     & 09.87                     & 13.24                      & 50.64                     & 36.87                     & 51.97                      & 21.65                     & 14.54                           & 23.75                      & {\underline{14.65}}               & {\underline{14.12}}               & 19.24                           \\
                           & CauMVC                      & Ours                             & \textbf{25.78}            & \textbf{26.84}            & \textbf{26.56}             & \textbf{80.99}            & \textbf{77.06}            & \textbf{81.12}             & \textbf{33.89}            & \textbf{27.95}                  & \textbf{34.73}             & \textbf{15.39}            & \textbf{15.12}            & \textbf{23.61}                  \\ \hline
\end{tabular}}
\caption{Multi-view clustering performance on eight benchmark datasets (Part 1/2). The optimal results are marked in \textbf{bold}, and the sub-optimal values are \underline{underlined}.}
\label{com_res_1}
\end{table*}

\section{Experiments}

In this section, we implement experiments to validate the superiority of CauMVC by addressing the following raised questions $\textbf{(RQ)}$: \textbf{RQ1}: How does CauMVC compare with other leading deep multi-view clustering techniques in terms of performance? \textbf{RQ2}: In what ways do the essential components of CauMVC enhance multi-view clustering results? \textbf{RQ3}: What is the effect of hyper-parameters on the efficacy of CauMVC? \textbf{RQ4}: What clustering structures are identified by CauMVC?



\subsection{Datasets \& Metric}

To assess the effectiveness of our proposed CauMVC, we conduct comprehensive experiments on eight widely used datasets, including BBCSport, Reuters, Caltech101-7, UCI-digit, Movies, WebKB, SUNRGB-D, and STL-10. A summary of the key information regarding these datasets is presented in Tab.~\ref{DATASET_INFO}. Besides, we adopt the commonly used metrics to verify the model performance, including clustering accuracy (ACC)~\cite{CCGC,MGCN}, normalized mutual information (NMI)~\cite{graph_learner,CONVERT}, and purity (PUR)~\cite{MFLVC}.

\begin{table}[t]
\centering
\scalebox{0.9}{
\begin{tabular}{@{}clclclc@{}}
\toprule
Dataset      &  & Samples &  & Clusters &  & Views \\ \midrule
BBCSport     &  & 544     &  & 5        &  & 2     \\
Movies       &  & 617     &  & 17       &  & 2     \\
WebKB       &  & 1051     &  & 2       &  & 2     \\
Reuters      &  & 1200    &  & 6        &  & 5     \\
Caltech101-7 &  & 1400    &  & 7        &  & 5     \\
UCI-digit    &  & 2000    &  & 10        &  & 3     \\
SUNRGB-D    &  & 10335    &  & 45        &  & 3     \\
STL-10        &  & 13000   &  & 10       &  & 4     \\ \bottomrule
\end{tabular}}
\caption{Dataset summary in this paper.}
\label{DATASET_INFO} 
\vspace{-10pt}
\end{table}

\subsection{Experiment Setup}
Our experiments are conducted on the PyTorch platform utilizing an NVIDIA 3090 GPU. For all baseline methods, we replicate the results using the original source code and configurations provided by the authors. DealMVC~\cite{DealMVC} is employed as the pretraining model to obtain the clustering results. In the proposed method, the algorithm is optimized using the Adam optimizer~\cite{adam}. The learning rate is set to 3e-3 for the BBCSport, Caltech-7, UCI-digit, and STL-10 datasets, and 5e-3 for the Movies dataset. The training is conducted for 500 epochs with a batch size of 256 for all datasets. The trade-off hyper-parameters $\alpha$ and $\beta$ are consistently set to 1.0. Detailed hyper-parameter settings are presented in Tab.~\ref{hyper}.

\begin{table*}[]
\centering
\scalebox{0.8}{
\begin{tabular}{c|ccccccccc}
\hline
\textbf{}                                  & \textbf{Dataset} & \textbf{BBCSport} & \textbf{Movies} & \textbf{WebKB} & \textbf{Reuters} & \textbf{Caltech101-7} & \textbf{UCI-digit} & \textbf{SUNRGB-D} & \textbf{STL-10} \\ \hline
\multirow{3}{*}{\textbf{Statistics}}       & Samples          & 544               & 617             & 1051             & 1200             & 1400                  & 2000               & 10335             & 13000           \\
                                           & Clusters         & 5                 & 17              & 2              & 6                & 7                     & 10                 & 45              & 10              \\
                                           & Views            & 2                 & 2               & 2                 & 5                & 5                     & 3                  & 3                 & 4               \\ \hline
\multirow{3}{*}{\textbf{Hyper-parameters}} & $\alpha$                & 1.0               & 1.0               & 1.0             & 1.0              & 1.0                   & 1.0                & 1.0               & 1.0             \\
                                           & $\beta$                & 1.0               & 1.0               & 1.0             & 1.0              & 1.0                   & 1.0                & 1.0               & 1.0             \\
                                           & Learning Rate    & 0.003             & 0.005           & 0.003            & 0.003            & 0.003                 & 0.003              & 0.003            & 0.003           \\ \hline
\end{tabular}}
\caption{Statistics and hyper-parameter settings of eight benchmark datasets.}
\label{hyper}
\end{table*}

To show the effectiveness and generalization of our CauMVC, we implement performance comparison experiments for our proposed CauMVC and 14 baselines, which can be divided into two groups, i.e., nine compared multi-view clustering algorithms with fully aligned data (SDMVC~\cite{SDMVC}, CoMVC~\cite{SiMVC}, SiMVC~\cite{SiMVC}, SDSNE~\cite{SDSNE}, MFLVC~\cite{MFLVC}, DSMVC~\cite{DSMVC}, GCFAggMVC~\cite{GCFAgg}, DealMVC~\cite{DealMVC}, TGM-MVC~\cite{TGM-MVC}), five compared methods with partially aligned data (PVC~\cite{PVC}, MvCLN~\cite{MvCLN}, SURE~\cite{SURE}, MVC-UM~\cite{MVC-UM}, RMCNC~\cite{sun2024robust}).

\begin{table*}[]
\centering
\scalebox{0.8}{
\begin{tabular}{c|cc|ccc|ccc|ccc|ccc}
\hline
\multirow{2}{*}{Aligned} & \multicolumn{2}{c|}{Mthods}                                    & \multicolumn{3}{c|}{BBCSport}                                                      & \multicolumn{3}{c|}{Reuters}                                                             & \multicolumn{3}{c|}{Caltech101-7}                                                  & \multicolumn{3}{c}{WebKB}                                                                           \\ \cline{2-15} 
                         & \multicolumn{2}{c|}{Evaluation metrics}                        & ACC                       & NMI                       & PUR                        & ACC                       & NMI                             & PUR                        & ACC                       & NMI                       & PUR                        & ACC                             & NMI                             & PUR                             \\ \hline
\multirow{10}{*}{Fully}  & SDMVC                       & TKDE 2021                        & 51.29                     & 32.86                     & 57.23                      & 17.67                     & 14.73                           & 18.58                      & 44.21                     & 30.97                     & 47.93                      & 75.07                           & 16.30                           & 78.12                           \\
                         & CoMVC                       & CVPR 2021                        & 37.50                     & 09.82                     & 40.81                      & 32.25                     & 13.58                           & 33.83                      & 40.84                     & 28.66                     & 69.95                      & 66.03                           & 13.66                           & 78.12                           \\
                         & SiMVC                       & CVPR 2021                        & 37.87                     & 10.99                     & 40.07                      & 33.58                     & 10.36                           & 33.67                      & 75.03                     & 37.07                     & 77.54                      & 72.79                           & 10.31                           & 78.12                           \\
                         & SDSNE                       & AAAI 2022                        & 76.62                     & 57.50                     & 75.84                      & 23.25                     & 20.28                           & 27.00                      & 76.14                     & 74.52                     & 78.93                      & 76.19                           & 14.55                           & 76.19                           \\
                         & MFLVC                       & CVPR 2022                        & 40.99                     & 27.79                     & 61.46                      & 39.92                     & 20.01                           & 41.42                      & 80.40                     & 70.30                     & 80.40                      & 71.65                           & 15.95                           & 78.12                           \\
                         & DSMVC                       & CVPR 2022                        & 41.91                     & 14.67                     & 47.59                      & 43.83                     & 18.11                           & 45.00                      & 62.64                     & 47.34                     & 64.43                      & 64.41                           & 11.62                           & 78.12                           \\
                         & GCFAggMVC                   & CVPR 2023                        & 59.01                     & 39.27                     & 66.91                      & 38.83                     & 21.47                           & 39.92                      & 83.36                     & 73.31                     & 83.66                      & 74.65                           & 15.47                           & 78.99                           \\
                         & DealMVC                     & ACM MM 2023                      & {\underline{80.70}}               & \textbf{65.59}            & {\underline{80.70}}                & {\underline{47.05}}               & 26.32                           & {\underline{48.36}}                & {\underline{88.71}}               & {\underline{80.95}}               & {\underline{88.71}}                & 79.26                           & 15.32                           & 79.26                           \\
                         & \multicolumn{1}{l}{TGM-MVC} & \multicolumn{1}{l|}{ACM MM 2024} & \multicolumn{1}{l}{39.15} & \multicolumn{1}{l}{17.92} & \multicolumn{1}{l|}{39.89} & \multicolumn{1}{l}{42.42} & \multicolumn{1}{l}{{\underline{29.65}}} & \multicolumn{1}{l|}{42.91} & \multicolumn{1}{l}{86.44} & \multicolumn{1}{l}{78.67} & \multicolumn{1}{l|}{83.64} & \multicolumn{1}{l}{{\underline{79.64}}} & \multicolumn{1}{l}{{\underline{16.61}}} & \multicolumn{1}{l}{{\underline{79.64}}} \\
                         & CauMVC                      & Ours                             & \textbf{83.09}            & {\underline{64.31}}               & \textbf{83.09}             & \textbf{49.08}            & \textbf{30.43}                  & \textbf{52.08}             & \textbf{90.43}            & \textbf{83.25}            & \textbf{90.43}             & \textbf{81.44}                  & \textbf{22.37}                  & \textbf{81.44}                  \\ \hline
\multirow{6}{*}{Partially} & PVC                         & NeurIPS 2020                     & 35.66                     & 00.36                     & 35.66                      & 16.83                     & 00.04                           & 16.83                      & 46.00                     & 29.15                     & 44.83                      & 54.12                           & 12.47                           & 68.12                           \\
                         & MvGLN                       & CVPR 2021                        & 32.39                     & 04.80                     & {\underline{44.82}}                & {\underline{32.84}}               & {{11.82}}                     & {\underline{35.32}}                & {\underline{48.97}}               & 35.77                     & 45.12                      & 54.70                           & {\underline{12.74}}                     & 73.70                           \\
                         & MVC-UM                      & KDD 2021                         & {\underline{35.85}}               & {\underline{14.75}}               & 36.21                      & 17.00                     & 00.82                           & 17.08                      & 41.52                     & {\underline{47.52}}                     & \textbf{83.39}             & {{52.15}}                     & 11.34                           & {{70.21}}                     \\
                         & SURE                        & TPAMI 2023                       & 30.96                     & 01.94                     & 36.03                      & 26.53                     & 04.91                           & 27.80                      & 32.31                     & {{22.57}}               & {34.20}             & {\underline{55.85}}                           & 10.15                           & {\underline{74.12}}                           \\
                         & RMCNC                       & TKDE 2024                        & 27.57                     & 11.13                     & 27.65                      & 21.45                     & {\underline{18.54}}                           & 23.49                      & 42.87                     & 34.51                     & 43.46                      & 50.24                           & 11.58                           & 73.09                           \\
                         & CauMVC                      & Ours                             & \textbf{56.64}            & \textbf{36.87}            & \textbf{56.64}             & \textbf{43.36}            & \textbf{21.23}                  & \textbf{46.09}             & \textbf{70.90}            & \textbf{54.54}            & {\underline{70.90}}                & \textbf{57.62}                  & \textbf{14.37}                  & \textbf{75.98}                  \\ \hline
\end{tabular}}
\caption{Multi-view clustering performance on eight benchmark datasets (Part 2/2). The optimal results are marked in \textbf{bold}, and the sub-optimal values are \underline{underlined}.}
\label{com_res_2}
\end{table*}

\begin{figure*}[t]
\centering
\begin{minipage}{0.23\linewidth}
\centerline{\includegraphics[width=1\textwidth]{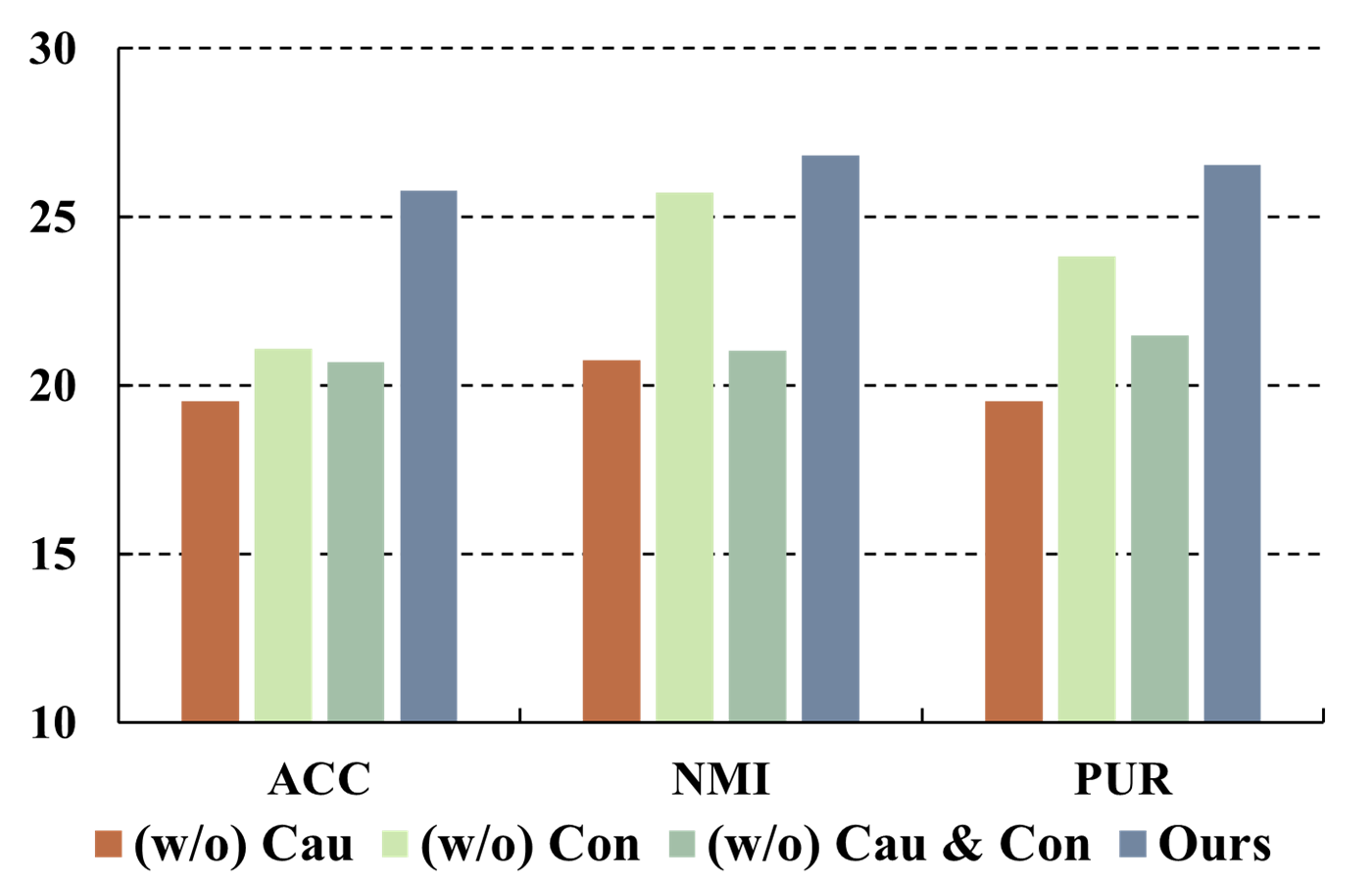}}
\centerline{{Movies}}
\end{minipage}
\begin{minipage}{0.23\linewidth}
\centerline{\includegraphics[width=1\textwidth]{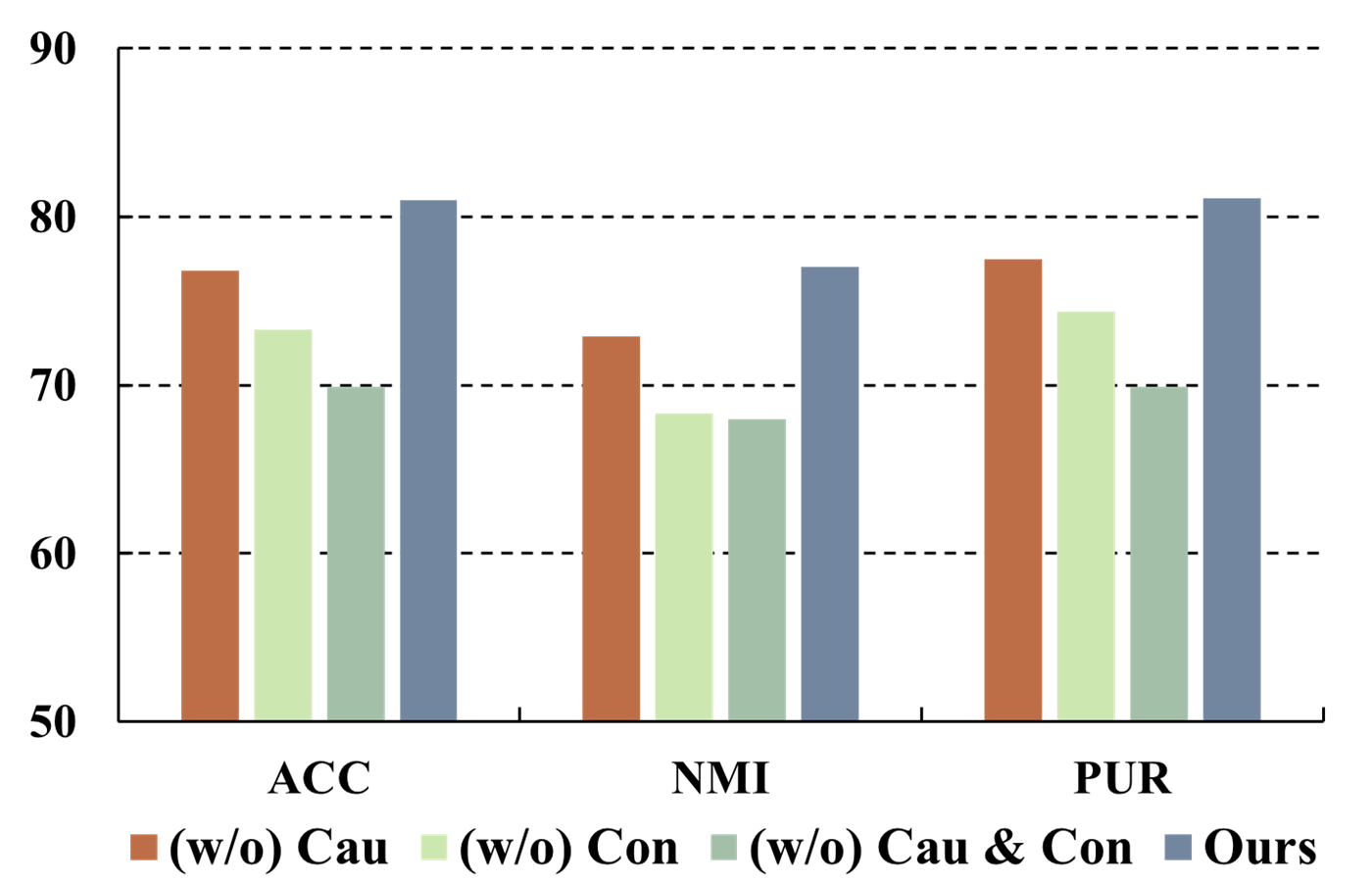}}
\centerline{{UCI-digit}}
\end{minipage}
\begin{minipage}{0.23\linewidth}
\centerline{\includegraphics[width=1\textwidth]{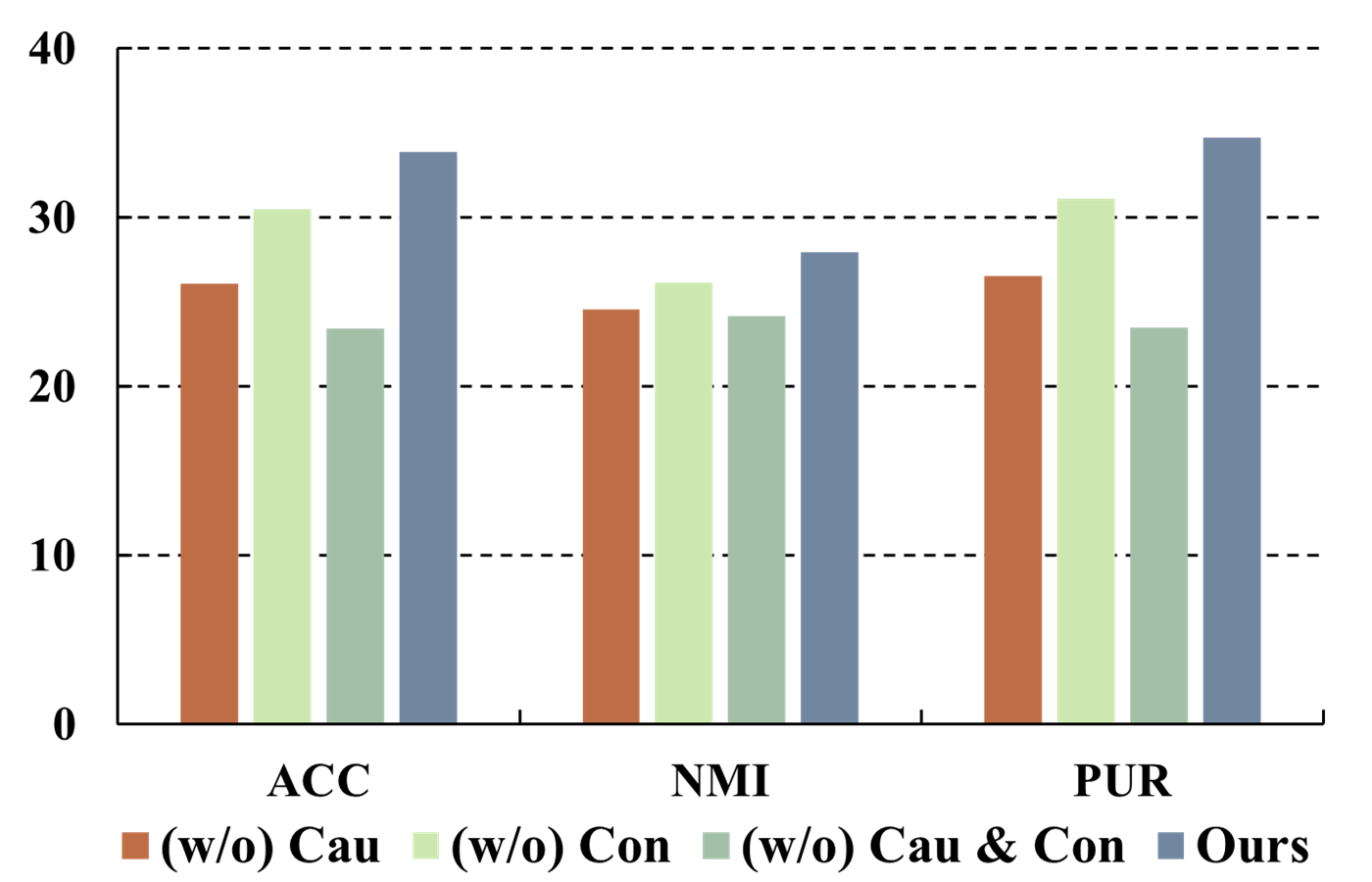}}
\centerline{{STL-10}}
\end{minipage}
\begin{minipage}{0.23\linewidth}
\centerline{\includegraphics[width=1\textwidth]{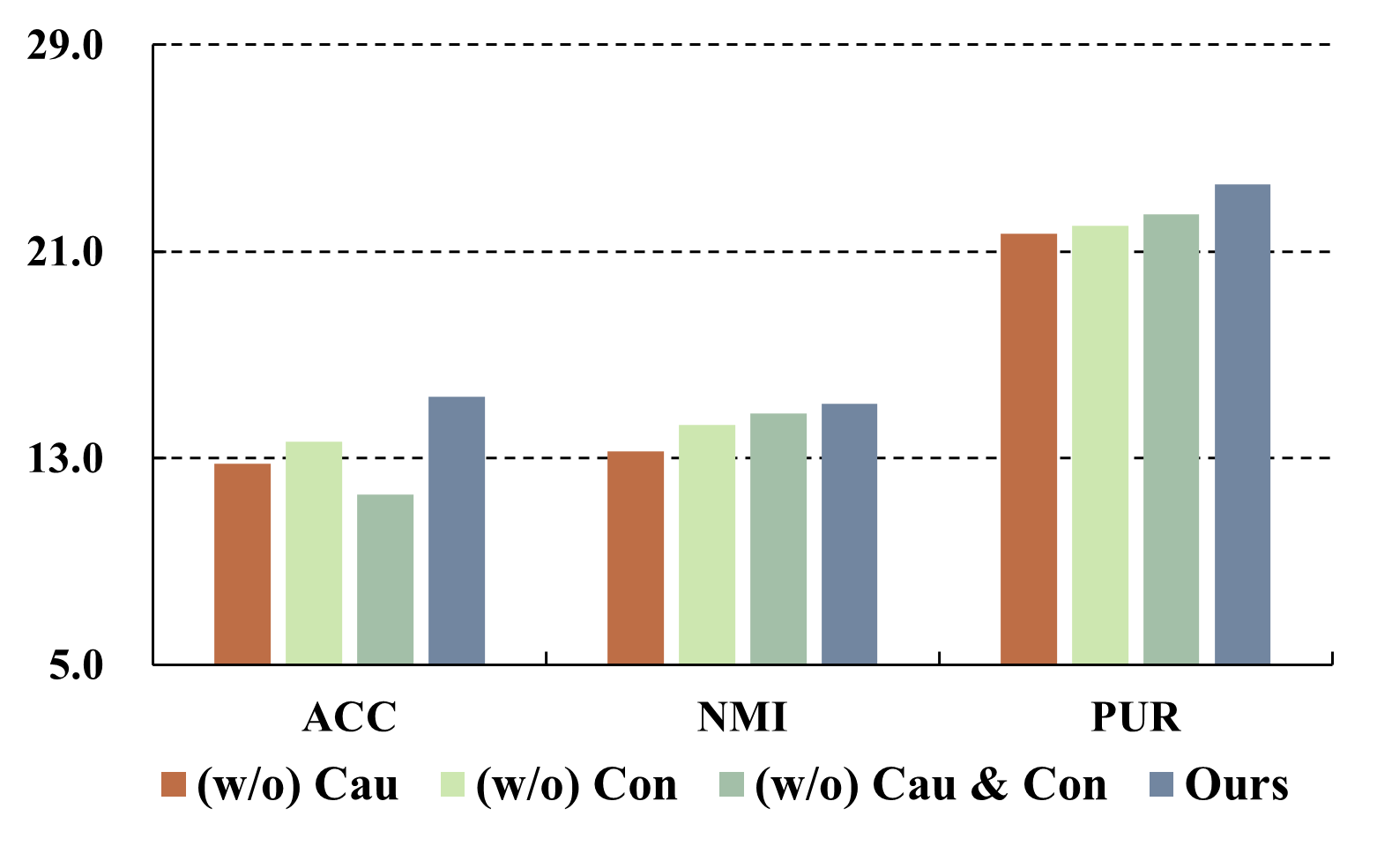}}
\centerline{{SUNRGB-D}}
\end{minipage}
\caption{Ablation studies on four datasets with partially aligned data. ``(w/o) Cau'', ``(w/o) Con'', ``(w/o) Cau\&Con'', and  ``Ours'' correspond to reduced models by individually removing the causal module, the contrastive regularizer, and all aforementioned modules combined, respectively.}
\label{ABLATION_MODULE}
\end{figure*}

\subsection{Performance Comparison~\textbf{(RQ1)}}
To evaluate the generalization capability of CauMVC, we conducted experiments on both aligned and partially aligned datasets, comparing the results against 11 baseline methods. Firstly, we construct the partially aligned data based on the eight datasets. To be concrete, following PVC~\cite{PVC}, we randomly spit the data into two partitions $\{\textbf{x}_{va}^{(v)}, \textbf{x}_{va}^{'(v)}\}_{v=1}^V$ with 50\% ratio.

\begin{figure*}[t]
\centering
\begin{minipage}{0.23\linewidth}
\centerline{\includegraphics[width=1\textwidth]{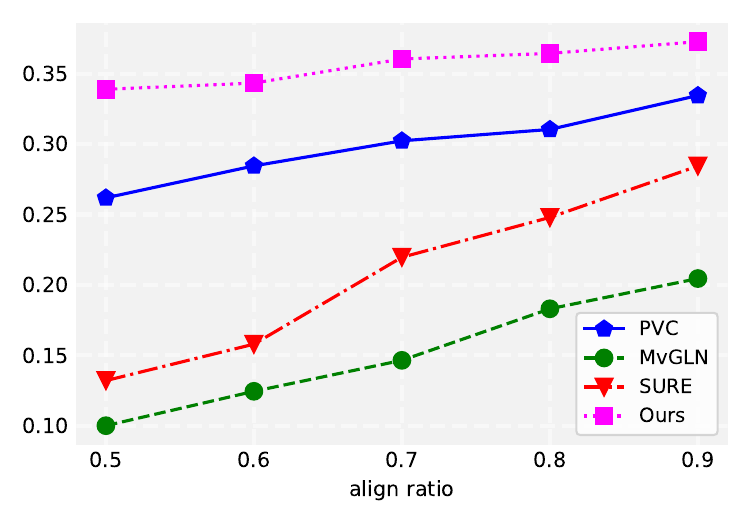}}
\centerline{{STL-10-ACC}}
\end{minipage}
\begin{minipage}{0.23\linewidth}
\centerline{\includegraphics[width=1\textwidth]{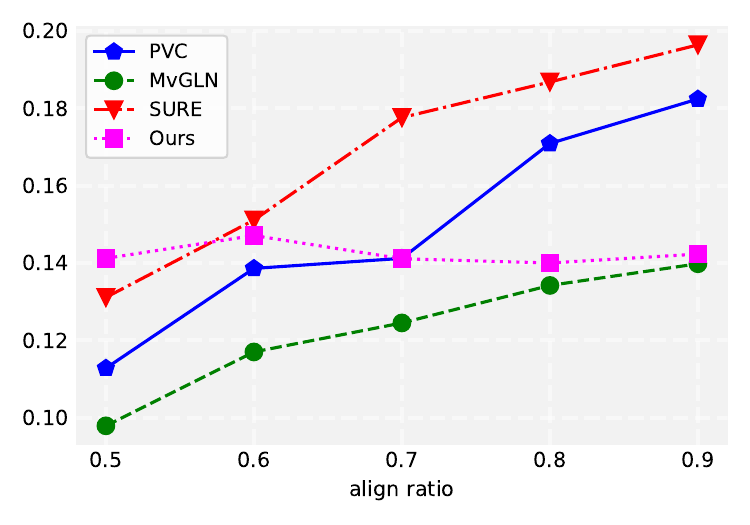}}
\centerline{{SUNRGB-D-NMI}}
\end{minipage}
\begin{minipage}{0.23\linewidth}
\centerline{\includegraphics[width=1\textwidth]{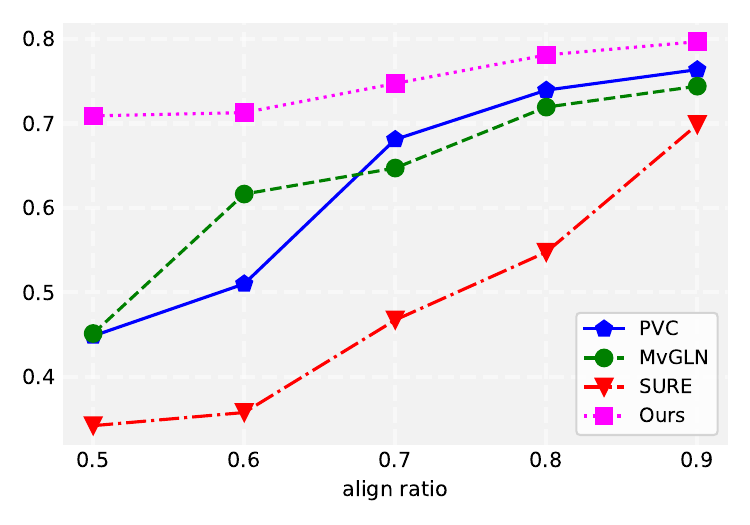}}
\centerline{{Caltech101-7-PUR}}
\end{minipage}
\begin{minipage}{0.23\linewidth}
\centerline{\includegraphics[width=1\textwidth]{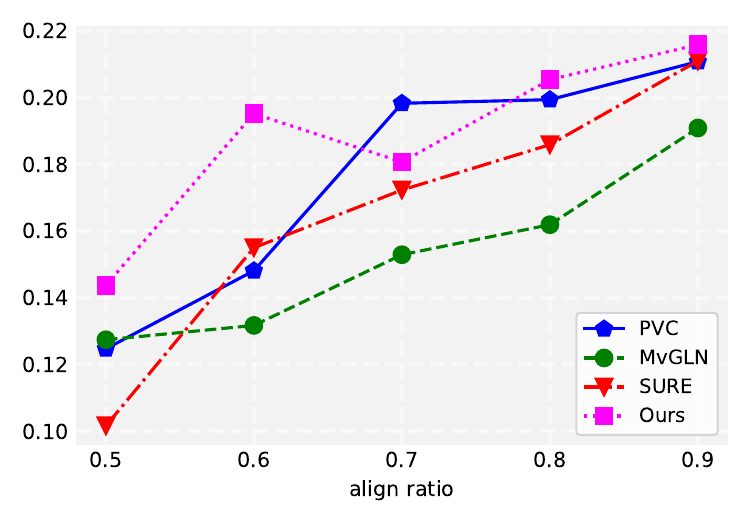}}
\centerline{{WebKB-10-NMI}}
\end{minipage}
\caption{Clustering performance on four Datasets with different aligned ratios.}
\label{align_acc}
\end{figure*}

\begin{figure}
\centering
\begin{minipage}{0.45\linewidth}
\centerline{\includegraphics[width=1\textwidth]{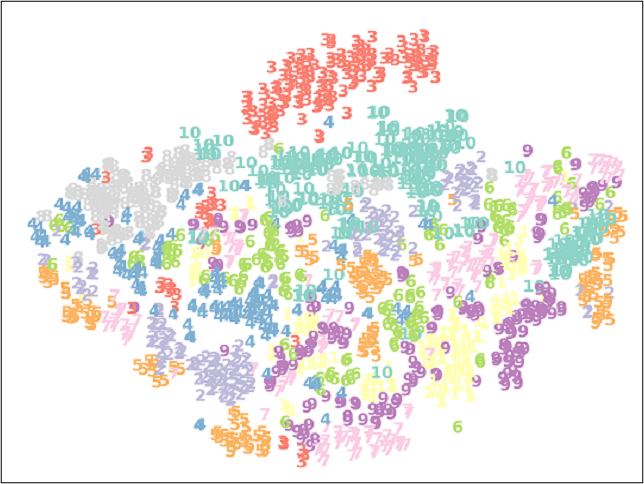}}
\vspace{5pt}
\centerline{{0 Epoch}}
\end{minipage}
\begin{minipage}{0.45\linewidth}
\centerline{\includegraphics[width=1\textwidth]{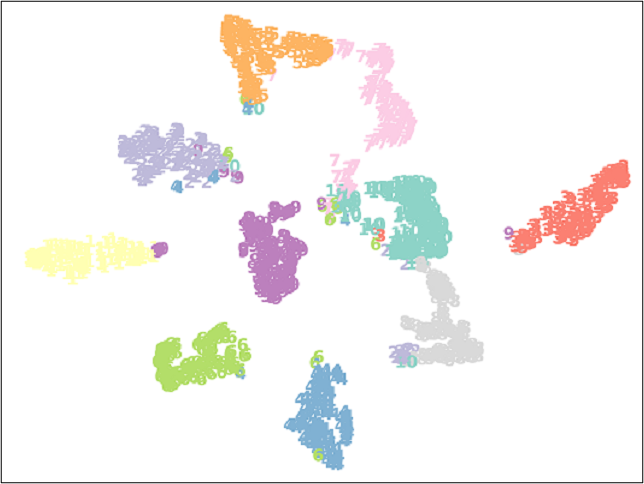}}
\vspace{5pt}
\centerline{{200 Epoch}}
\end{minipage}
\caption{Visualization of the representations during the training process on UCI-digit dataset.}
\label{t_sne}
\vspace{-10pt}
\end{figure}

The comparative experimental results are shown in Tab.~\ref{com_res_1} and Tab.~\ref{com_res_2}. We highlight the optimal results in bold, and the sub-optimal results with an underline. From those results, we have the following observations: (1) In the fully aligned data scenario, our CauMVC outperforms other multi-view clustering methods on eight datasets. Taking the results on the Caltech101-7 dataset as an example, CauMVC surpasses the second-best multi-view clustering method by 1.72\%, 2.30\%, and 1.72\% in ACC, NMI, and PUR, respectively. (2) CauMVC demonstrates promising clustering performance when dealing with partially aligned data. We attribute this to CauMVC's use of causal modeling in the multi-view clustering process and its ability to estimate the post-intervention probability. (3) CauMVC achieves superior performance in both fully aligned data and partially aligned data scenarios. The high performance showcases the strong generalization ability of CauMVC under data shifts, from fully aligned data to partially aligned data. In summary, the above observations confirm the remarkable performance of our proposed CauMVC.

    


    

\textbf{Influence of Different Aligned Ratios}: Moreover, we perform additional experiments to investigate the model performance across various aligned ratios on eight datasets. The aligned ratio ranges from 0.5 to 0.9. The results for ACC, NMI, and PUR are illustrated in Fig.\ref{align_acc}. It is evident that CauMVC surpasses other baseline models across different aligned ratios in most scenarios. This indicates the robust generalization capability of CauMVC in managing partially aligned data.


\subsection{Ablation Studies~\textbf{(RQ2)}}

In this subsection, we conduct ablation studies with partially aligned data to verify the effectiveness of our designed modules, i.e, causal module and contrastive regularizer. Concretely, ``(w/o) Cau'', ``(w/o) Con'', ``(w/o) Cau\&Con'', and  ``Ours'' represent the ablated models where the causal module, the contrastive regularizer, and both modules combined, respectively, are individually removed. In the ``(w/o) Cau\&Con'' configuration, we employ an autoencoder network as the backbone to derive representations for the downstream clustering task. The corresponding results are illustrated in Fig.~\ref{ABLATION_MODULE}. The findings indicate that omitting any of the proposed modules results in a significant reduction in clustering performance, highlighting the critical contribution of each designed module to the overall efficacy. We further delve into the underlying reasons as follows: (1) We resort the multi-view clustering from the causal perspective. The model generalization is improved when the input is partially aligned data, thus achieving promising performance. (2) The contrastive module could push the positive sample close, and pull the negative sample away, enhancing the model's discriminative capacity. The model could achieve better clustering outcomes. Besides, we perform ablation studies with fully aligned data to assess the effectiveness of our designed modules, namely the causal module and contrastive regularizer. The results in Fig.~2 of the Appendix clearly demonstrate that eliminating any of the designed modules results in a significant drop in clustering performance. This underscores the crucial role each module plays in achieving optimal overall performance.


\subsection{Hyper-parameter Analysis~\textbf{(RQ3)}}


To further examine the impact of the parameters $\alpha$ and $\beta$ on our model, we conducted experiments using UCI-digit dataset. In particular, we analyzed parameter values within the range of $\{0.01, 0.1, 1.0, 10, 100\}$. According to the results presented in Fig.~\ref{sen_alpha}. We could find the following observations. (1) When $\alpha$ and $\beta$ are assigned extreme values (0.1 or 100), the clustering performance tends to decline. We speculate that this decline is due to the disruption of the loss function balance. Additionally, the model demonstrates optimal performance when the trade-off parameters are approximately $1.0$. (2) From the results, we find that $\alpha$ has a more pronounced effect on model performance, suggesting that the causal model greatly enhances the overall efficacy of the model.





\subsection{Visualization Experiment~\textbf{(RQ4)}}


In this subsection, we utilize visualization method to intuitively demonstrate the effectiveness of CauMVC. Specifically, we use the $t$-SNE~\cite{TSNE} algorithm as the baseline to visualize the distribution of the learned embeddings of CauMVC on the UCI-digit dataset. The experimental results are presented in Fig.\ref{t_sne}, the results indicate that, with the training procedure, CauMVC more effectively reveals the intrinsic clustering structure compared to the raw features. More detailed visualizations of the experiments are presented in Fig.6 in the Appendix.

\begin{figure}[t]
\centering
\begin{minipage}{0.32\linewidth}
\centerline{\includegraphics[width=\textwidth]{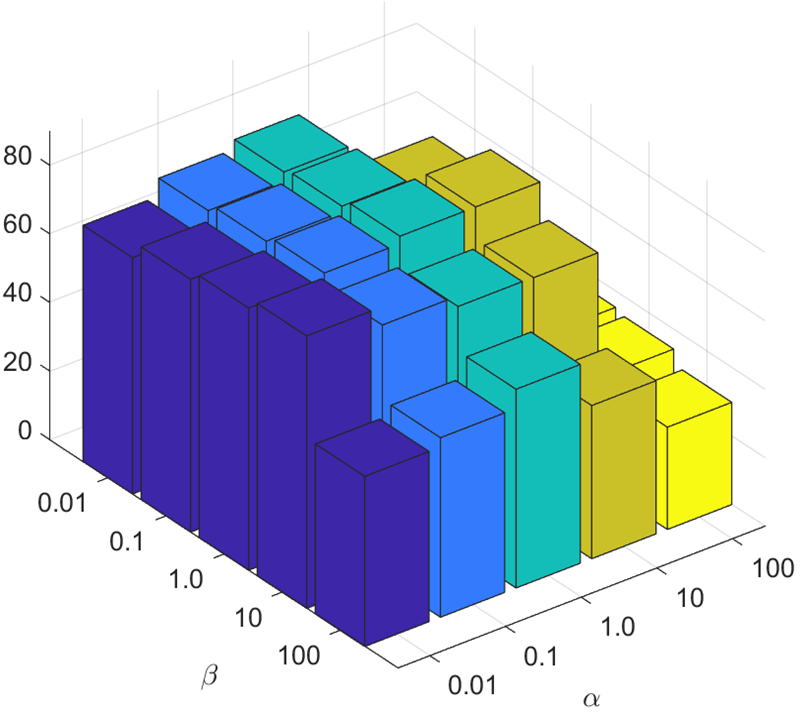}}
\vspace{3pt}
{\centerline{ACC}}
\end{minipage}
\begin{minipage}{0.32\linewidth}
\centerline{\includegraphics[width=\textwidth]{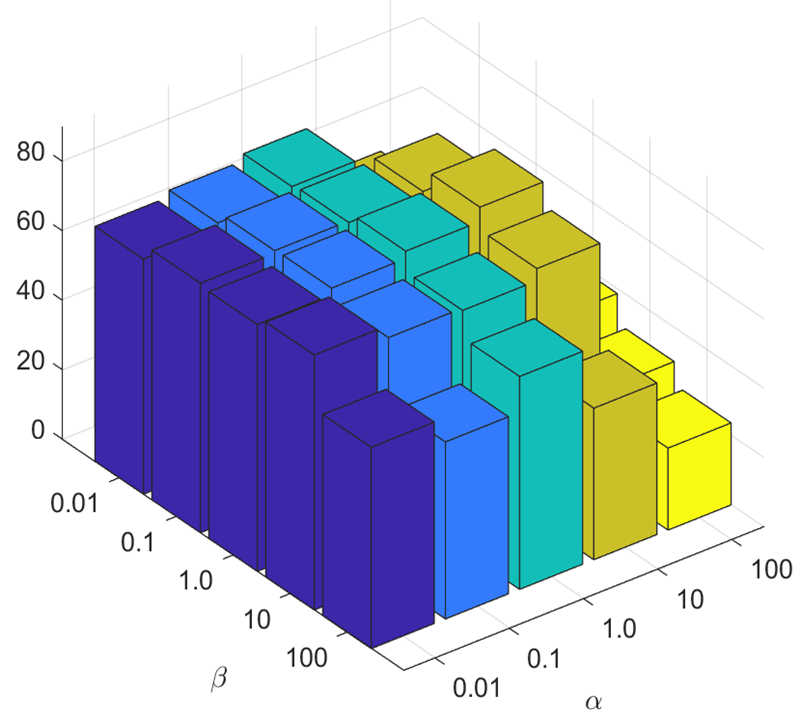}}
\vspace{3pt}
{\centerline{NMI}}
\end{minipage}
\begin{minipage}{0.32\linewidth}
\centerline{\includegraphics[width=\textwidth]{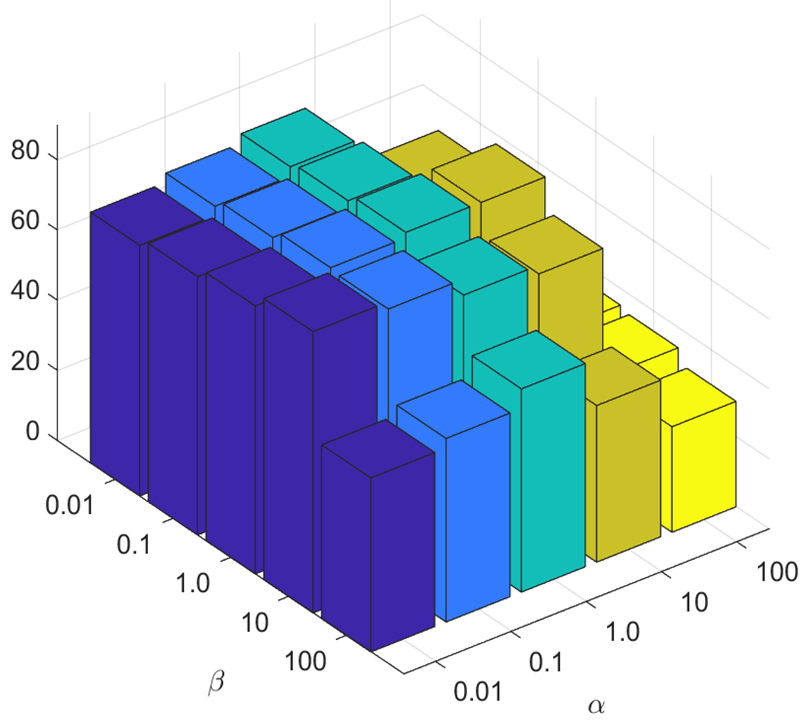}}
\vspace{3pt}
{\centerline{PUR}}
\end{minipage}
\caption{Sensitivity analysis of the hyper-parameter $\alpha$ and $\beta$ on UCI-digit datasets.}
\label{sen_alpha}
\vspace{-10pt}
\end{figure}


\section{Conclusion}

In this paper, we consider the model performance decreasing caused by data shift as a generalization problem. The objective of model is to achieve strong generalization on both fully and partially aligned data. We formulate this issue from a causal view, and design CauMVC to perform causal modeling of the procedure. Then, we conduct post-intervention inference to obtain the clustering results. Moreover, we design a contrastive regularizer to improve the discriminative capacity of the model. Extensive experiments on both fully aligned and partially aligned data have demonstrated the strong generalization and effectiveness of our CauMVC. This paper makes an initial attempt to explore model generalization through causal representation learning. In the future, how to deal with more challenging problem into a uniform causal framework is a interesting direction, such as incomplete data and noisy data.

\newpage

\section{Acknowledgments}
 This work is supported in by the National Science and Technology Innovation 2030 Major Project of China under Grant No. 2022ZD0209103, the Major Program Project of Xiangjiang Laboratory under Grant No. 24XJJCYJ01002, the National Science Fund for Distinguished Young Scholars of China under Grant No. 62325604, the National Natural Science Foundation of China (project No. 62406329, 62476280, 62276271, 62476281, 62406329), the National Natural Science Foundation of China Joint Found under Grant No. U24A20323, and the Program of China Scholarship Council (No. 202406110009).

\appendix

\begin{figure*}[t]
\centering
\small
\begin{minipage}{0.15\linewidth}
\centerline{\includegraphics[width=1\textwidth]{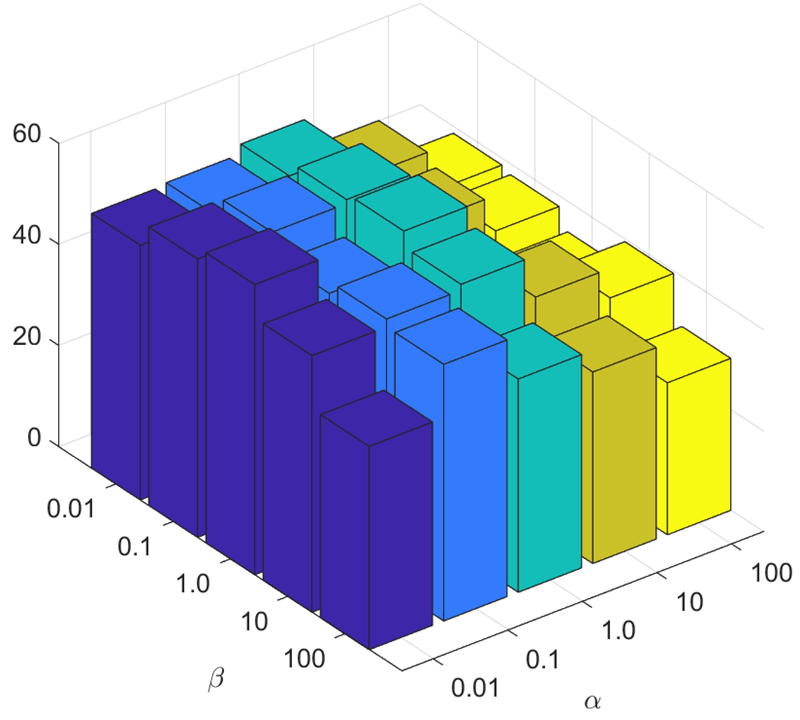}}
\vspace{3pt}
{\centerline{ACC}}
\end{minipage}
\begin{minipage}{0.15\linewidth}
\centerline{\includegraphics[width=\textwidth]{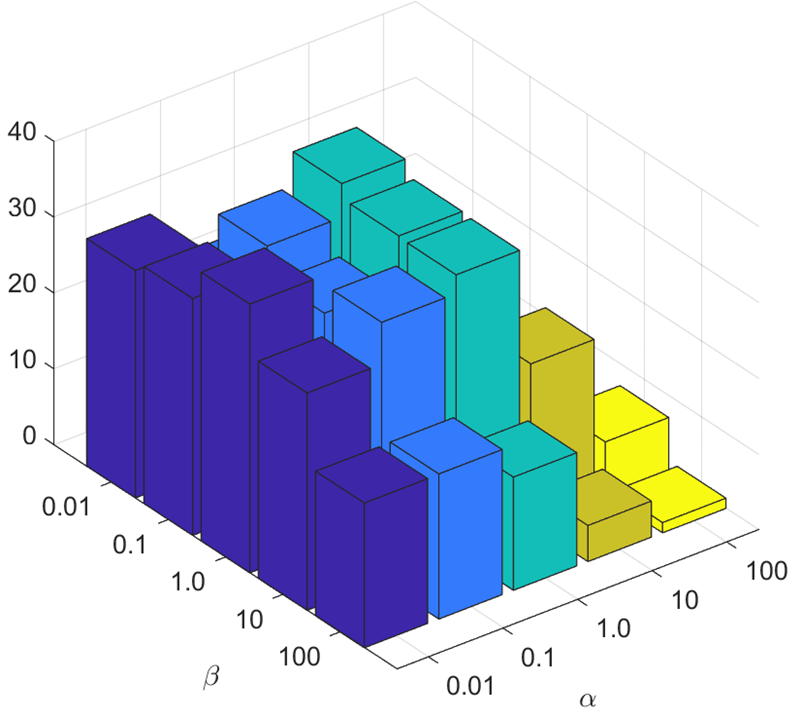}}
\vspace{3pt}
{\centerline{NMI}}
\end{minipage}
\begin{minipage}{0.15\linewidth}
\centerline{\includegraphics[width=\textwidth]{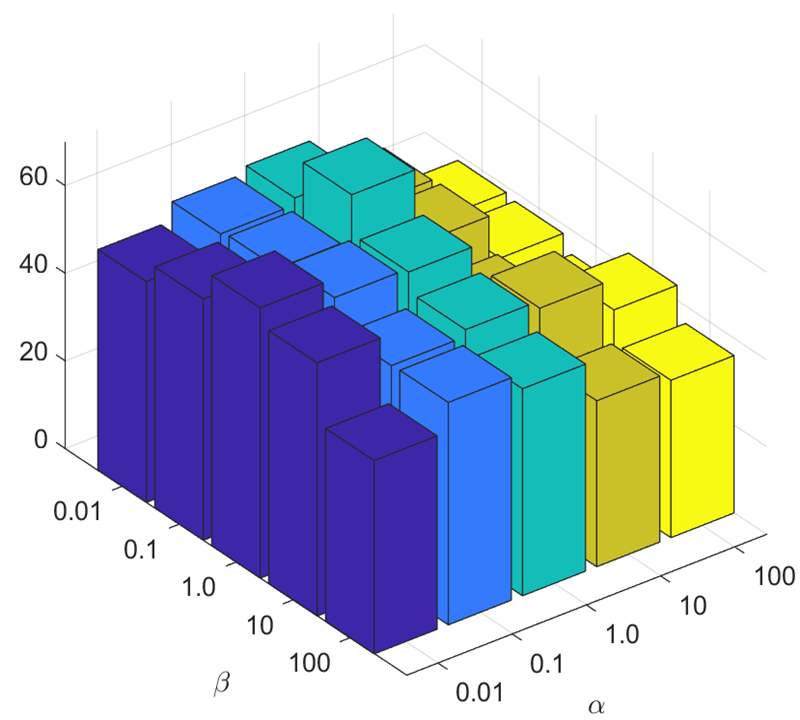}}
\vspace{3pt}
{\centerline{PUR}}
\end{minipage}
\begin{minipage}{0.15\linewidth}
\centerline{\includegraphics[width=\textwidth]{Figure/sen/uci_acc.png}}
\vspace{3pt}
{\centerline{ACC}}
\end{minipage}
\begin{minipage}{0.15\linewidth}
\centerline{\includegraphics[width=\textwidth]{Figure/sen/uci_nmi.png}}
\vspace{3pt}
{\centerline{NMI}}
\end{minipage}
\begin{minipage}{0.15\linewidth}
\centerline{\includegraphics[width=\textwidth]{Figure/sen/uci_pur.png}}
\vspace{3pt}
{\centerline{PUR}}
\end{minipage}
\caption{Sensitivity analysis of the hyper-parameter $\alpha$ and $\beta$ on BBCSport and UCI-digit datasets.}
\label{sen_alpha}
\end{figure*}

\begin{figure*}
\centering
\begin{minipage}{0.23\linewidth}
\centerline{\includegraphics[width=1\textwidth]{Figure/ablation/movies.png}}
\vspace{3pt}
\centerline{{Movies}}
\centerline{\includegraphics[width=1\textwidth]{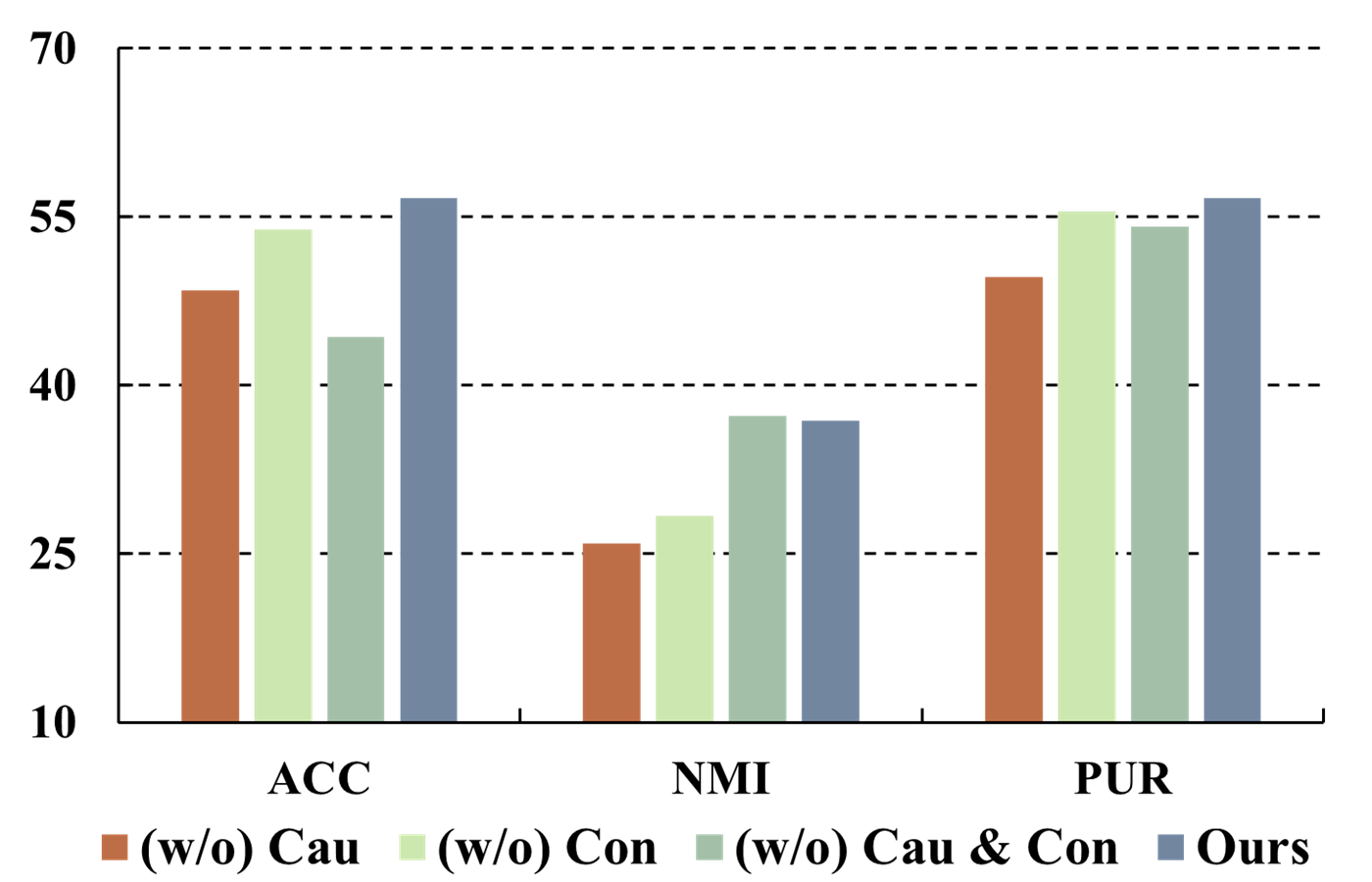}}
\vspace{3pt}
\centerline{{BBCSport}}
\end{minipage}
\begin{minipage}{0.23\linewidth}
\centerline{\includegraphics[width=1\textwidth]{Figure/ablation/uci-digit.png}}
\vspace{3pt}
\centerline{{UCI-digit}}
\centerline{\includegraphics[width=1\textwidth]{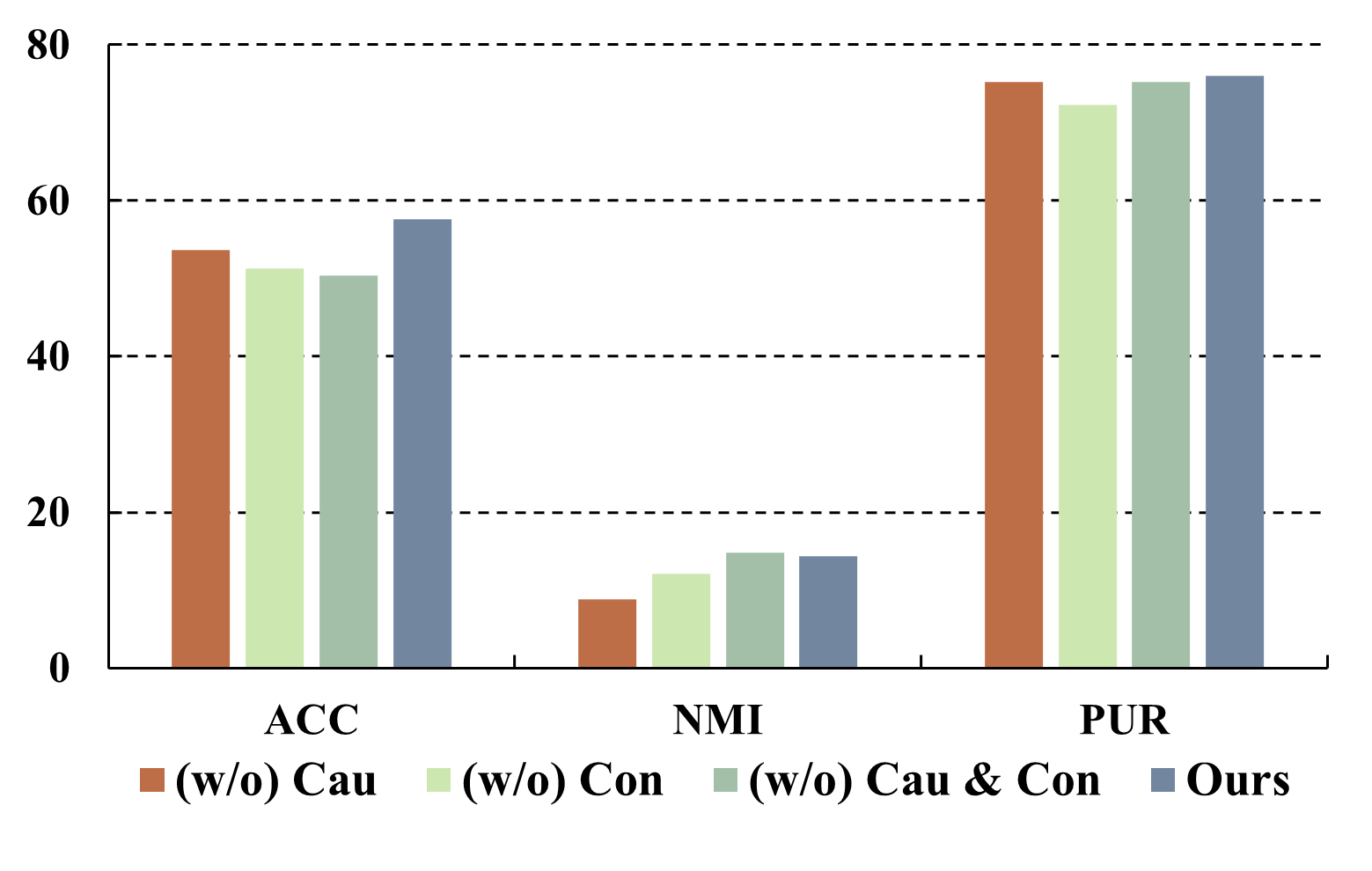}}
\vspace{3pt}
\centerline{{WebKB}}
\end{minipage}
\begin{minipage}{0.23\linewidth}
\centerline{\includegraphics[width=1\textwidth]{Figure/ablation/stl10.png}}
\vspace{3pt}
\centerline{{STL-10}}
\centerline{\includegraphics[width=1\textwidth]{Figure/ablation/sunrgb-d.png}}
\vspace{3pt}
\centerline{{SUNRGB-D}}
\end{minipage}
\begin{minipage}{0.23\linewidth}
\centerline{\includegraphics[width=1\textwidth]{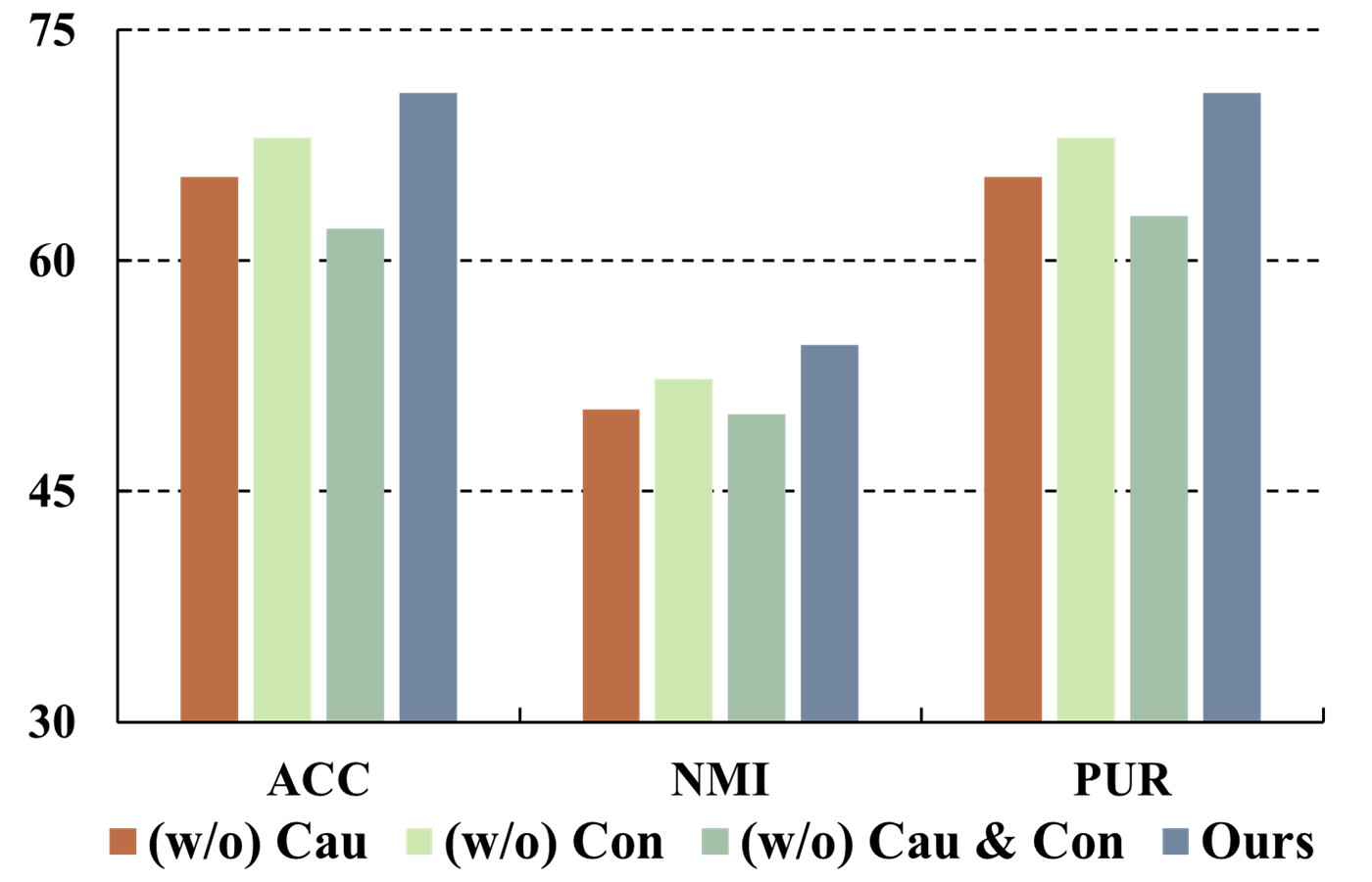}}
\vspace{3pt}
\centerline{{Caltech101-7}}
\centerline{\includegraphics[width=1\textwidth]{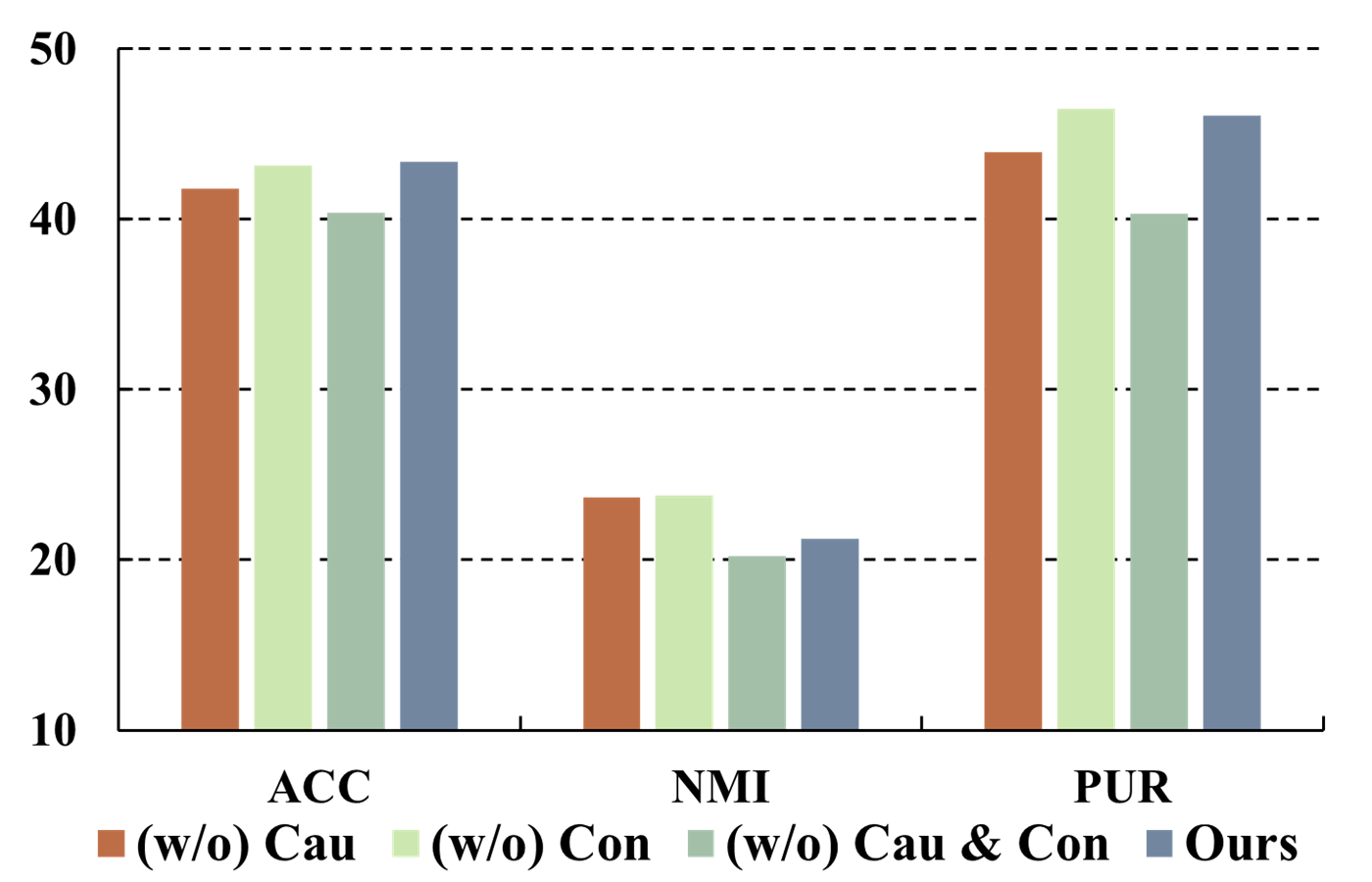}}
\vspace{3pt}
\centerline{{Reuters}}
\end{minipage}
\caption{Ablation studies on eight datasets with partially aligned data. ``(w/o) Cau'', ``(w/o) Con'', ``(w/o) Cau\&Con'', and  ``Ours'' correspond to reduced models by individually removing the causal module, the contrastive regularizer, and all aforementioned modules combined, respectively.}
\label{ABLATION_MODULE}
\end{figure*}

\begin{figure*}
\centering
\begin{minipage}{0.23\linewidth}
\centerline{\includegraphics[width=1\textwidth]{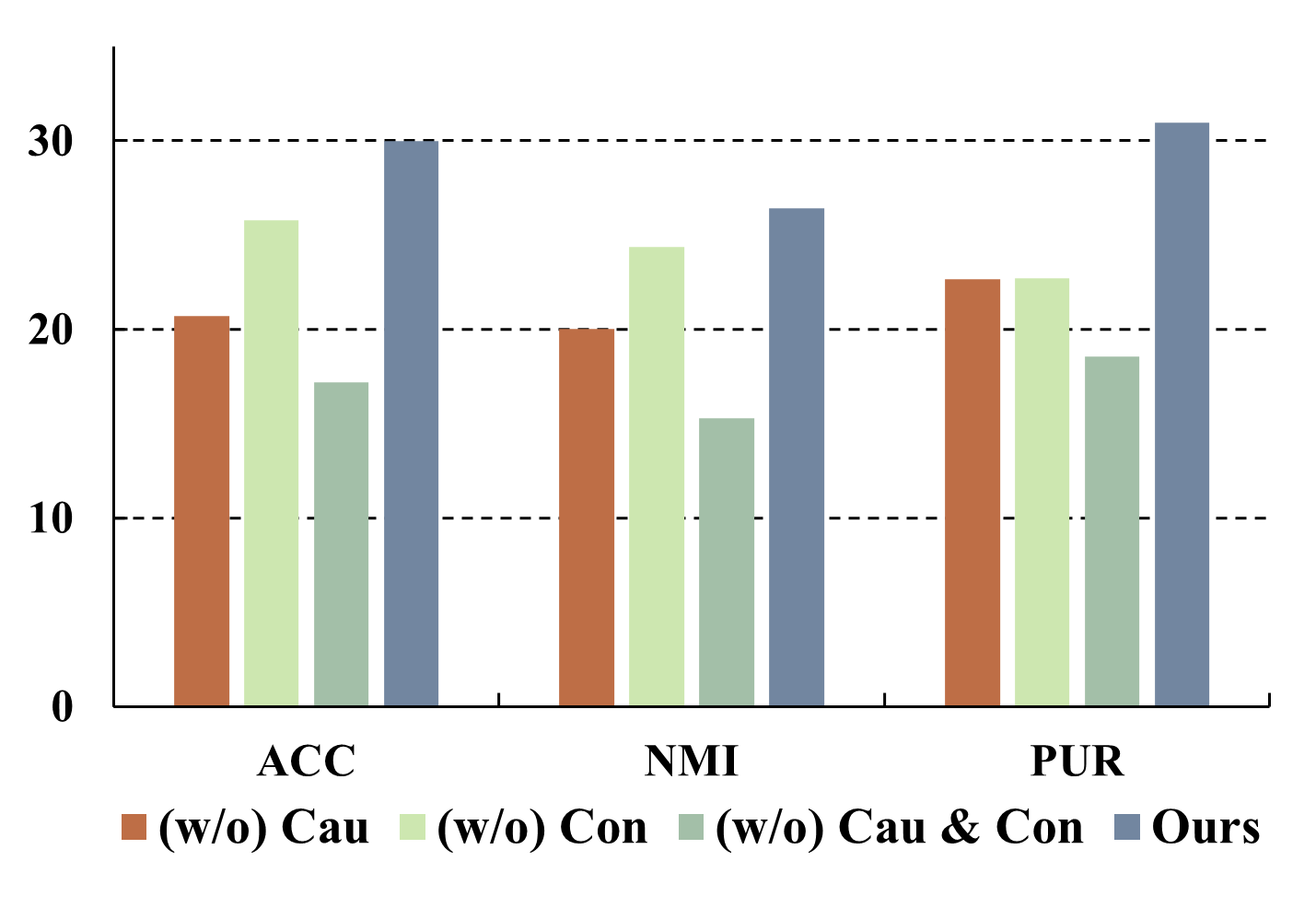}}
\vspace{3pt}
\centerline{{Movies}}
\centerline{\includegraphics[width=1\textwidth]{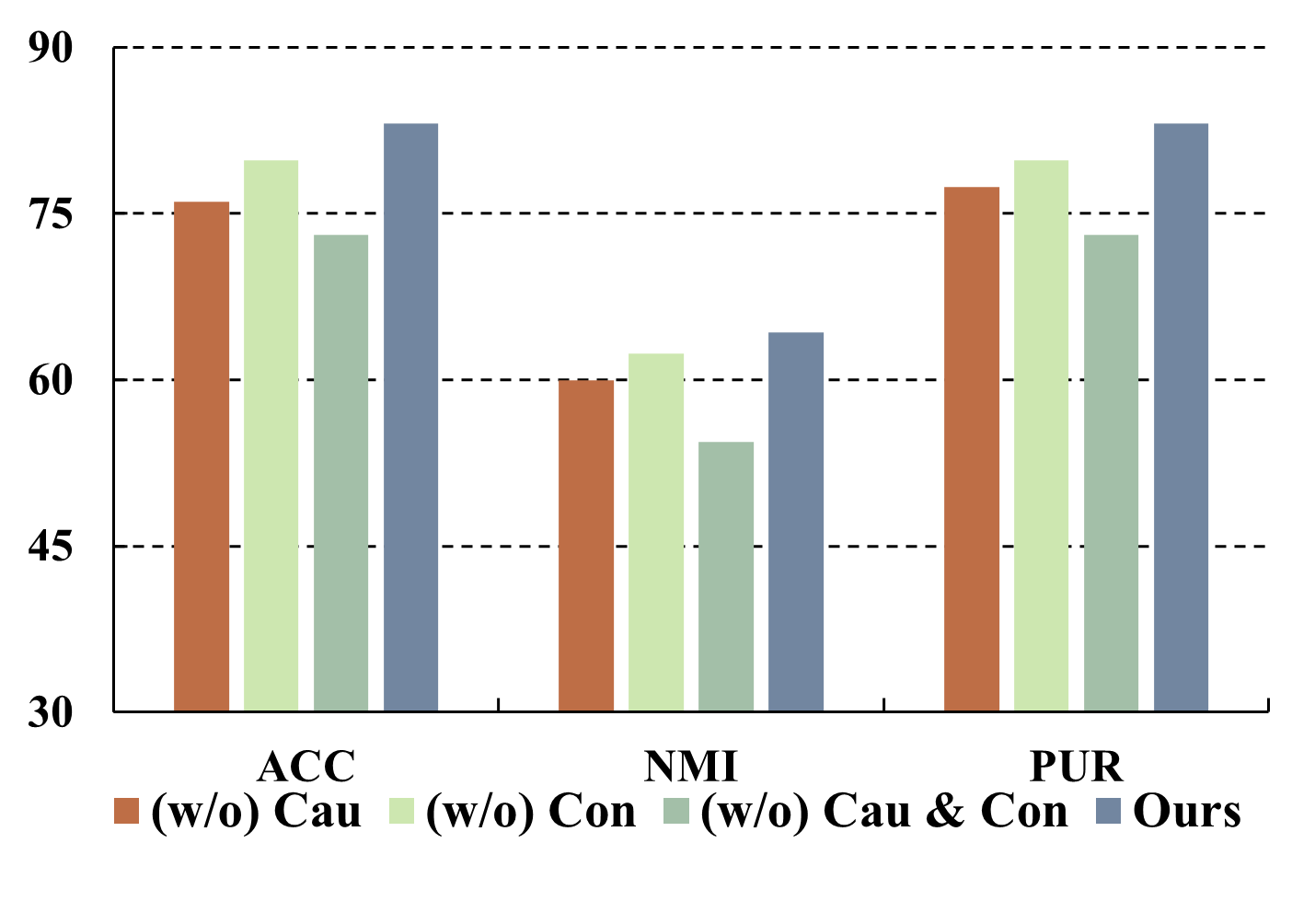}}
\vspace{3pt}
\centerline{{BBCSport}}
\end{minipage}
\begin{minipage}{0.23\linewidth}
\centerline{\includegraphics[width=1\textwidth]{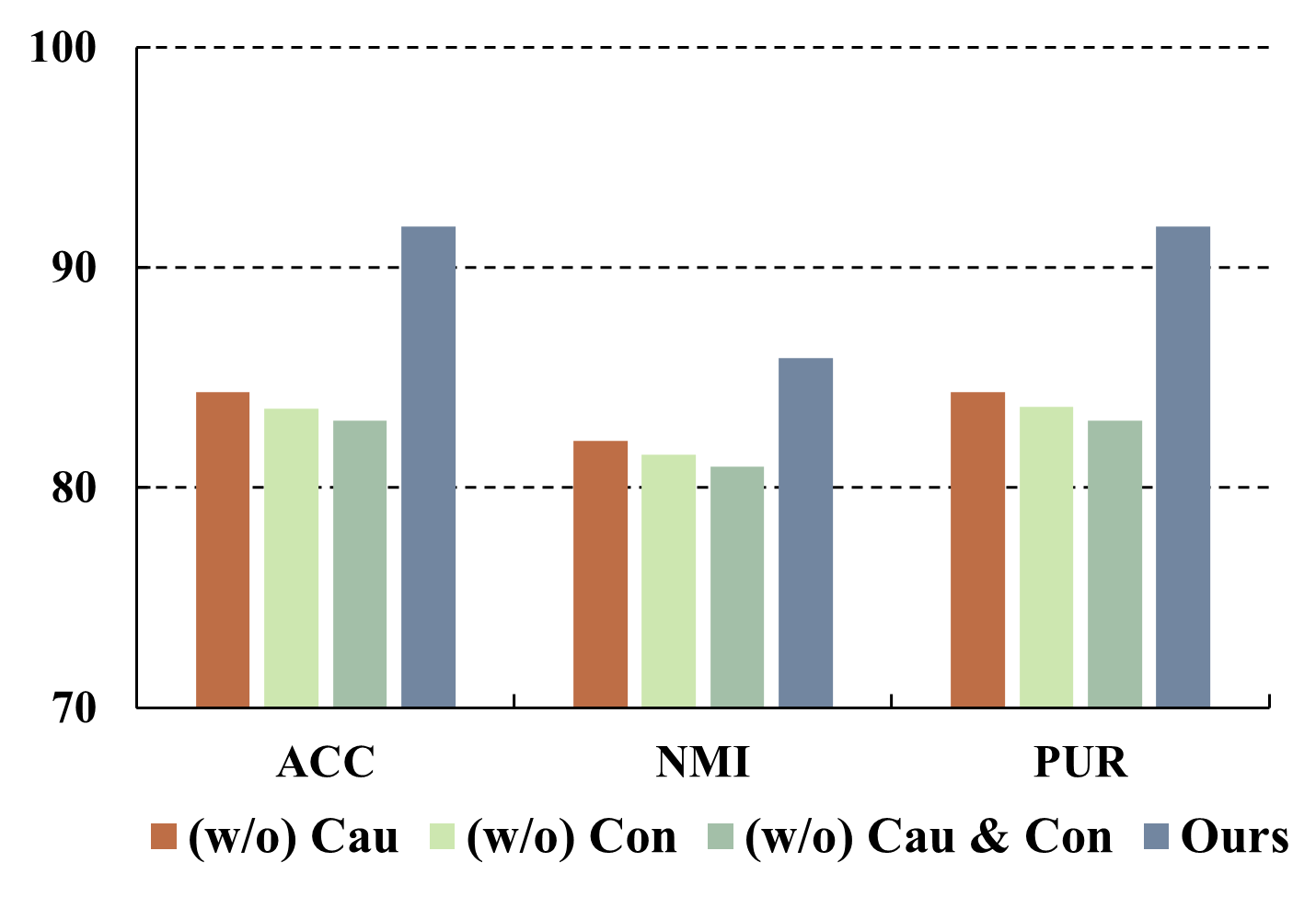}}
\vspace{3pt}
\centerline{{UCI-digit}}
\centerline{\includegraphics[width=1\textwidth]{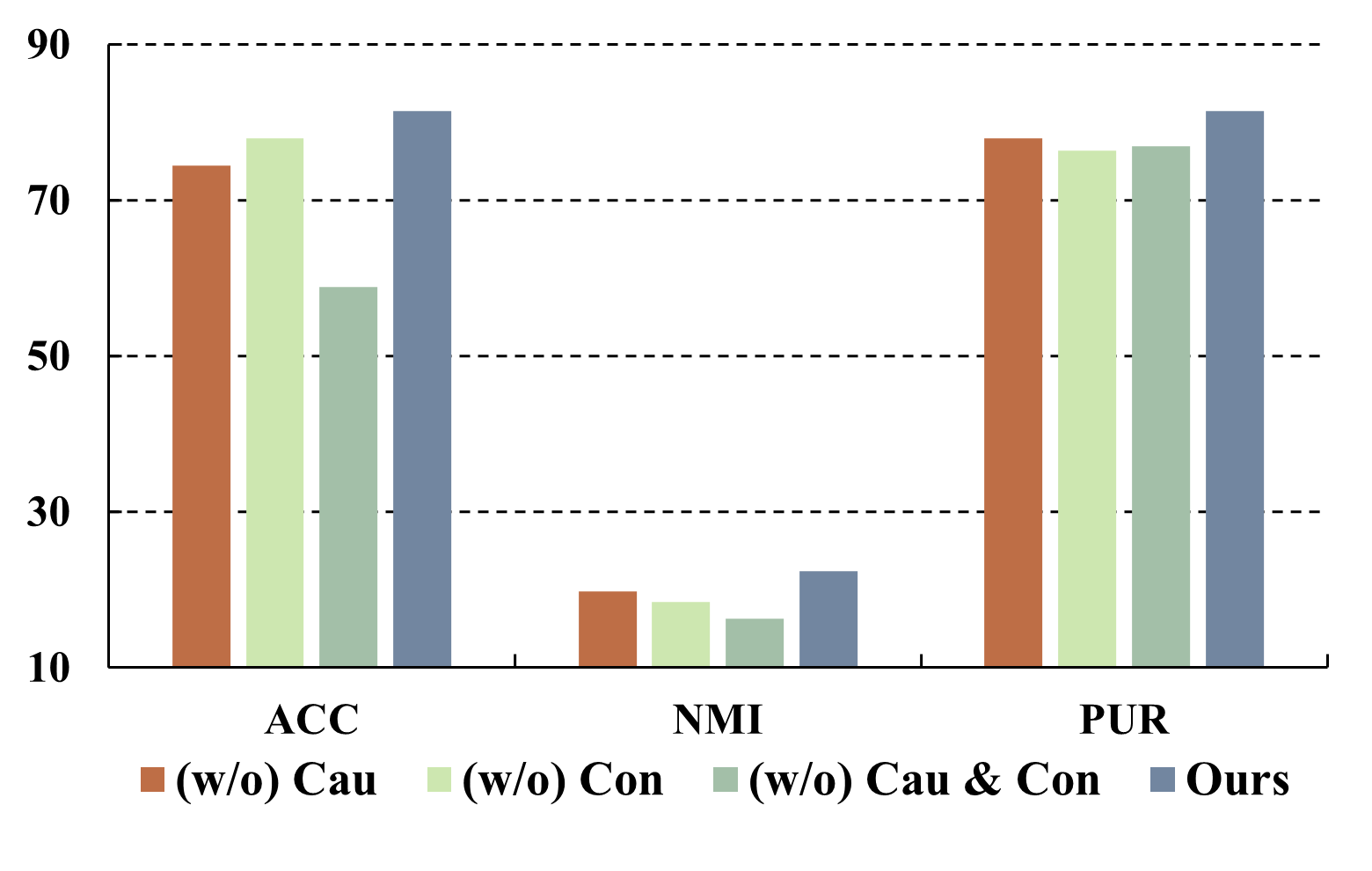}}
\vspace{3pt}
\centerline{{WebKB}}
\end{minipage}
\begin{minipage}{0.23\linewidth}
\centerline{\includegraphics[width=1\textwidth]{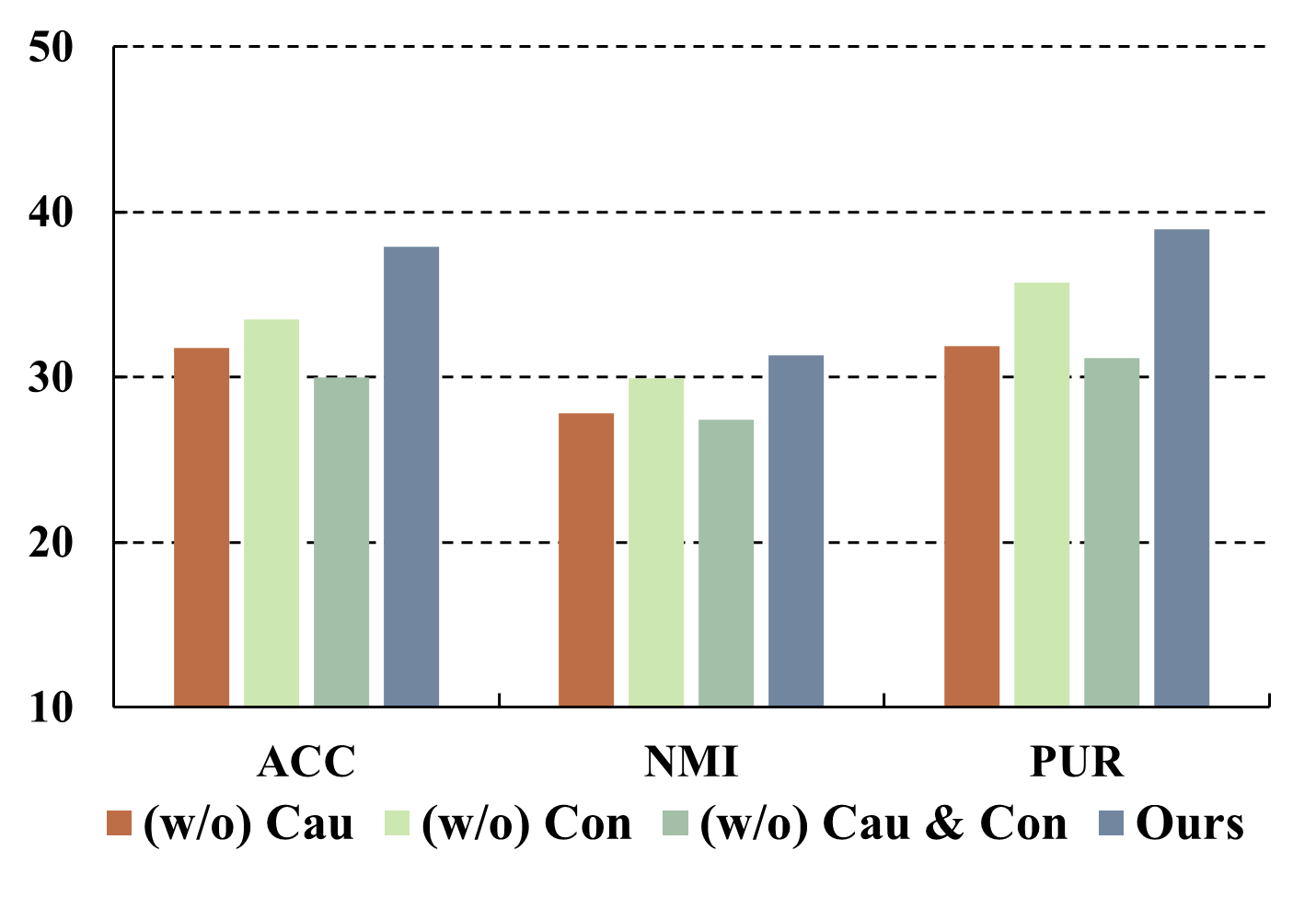}}
\vspace{3pt}
\centerline{{STL-10}}
\centerline{\includegraphics[width=1\textwidth]{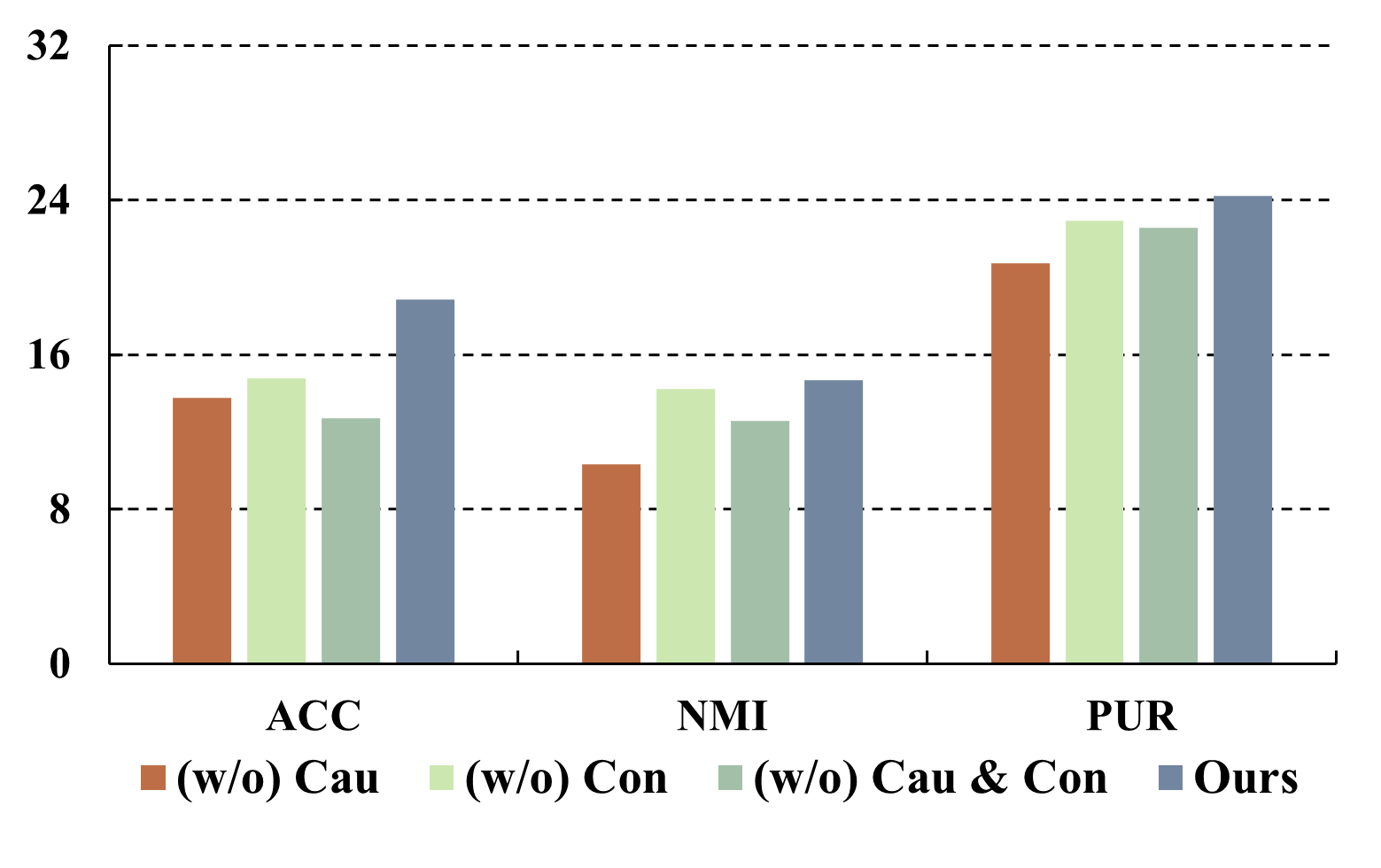}}
\vspace{3pt}
\centerline{{SUNRGB-D}}
\end{minipage}
\begin{minipage}{0.23\linewidth}
\centerline{\includegraphics[width=1\textwidth]{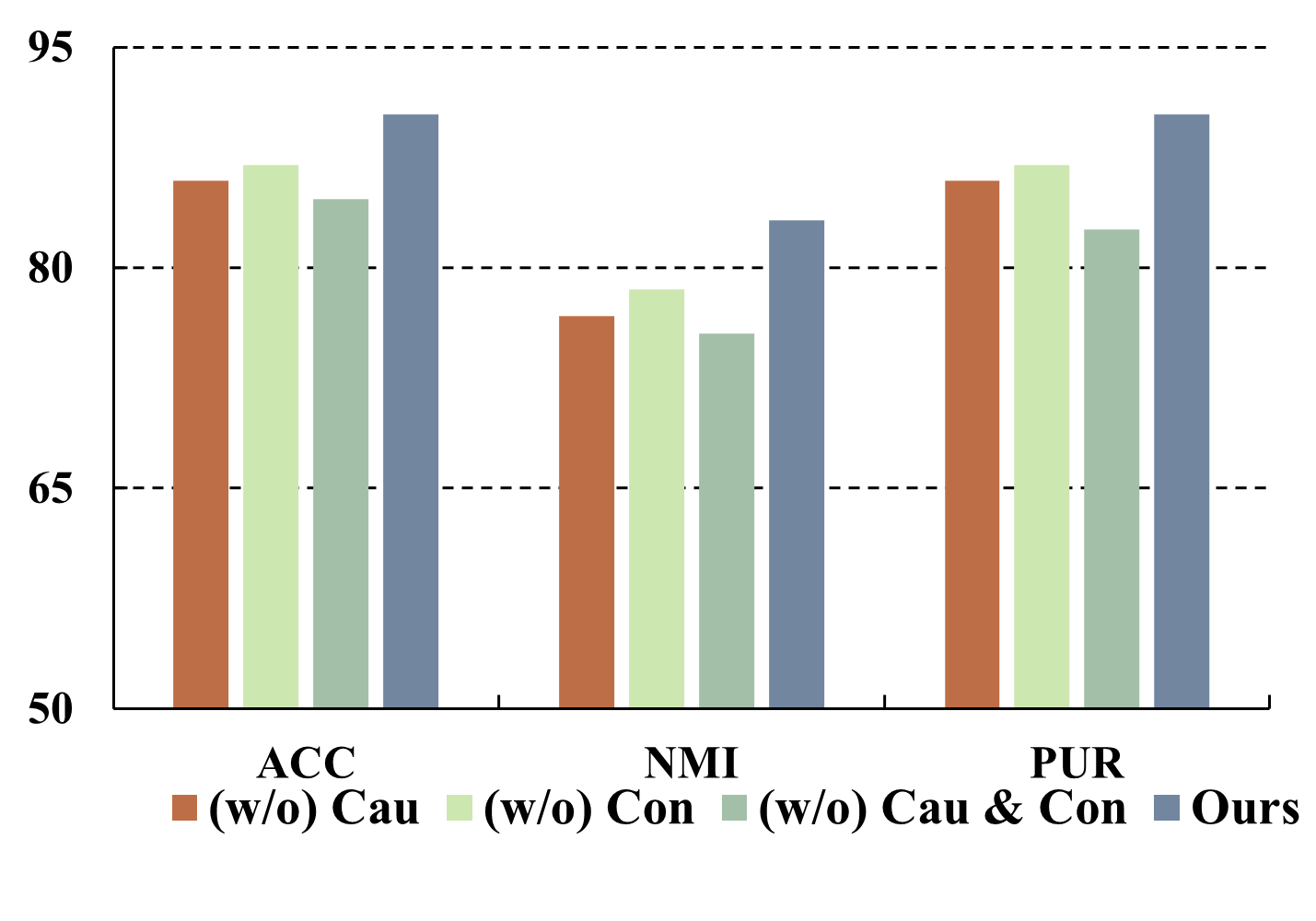}}
\vspace{3pt}
\centerline{{Caltech101-7}}
\centerline{\includegraphics[width=1\textwidth]{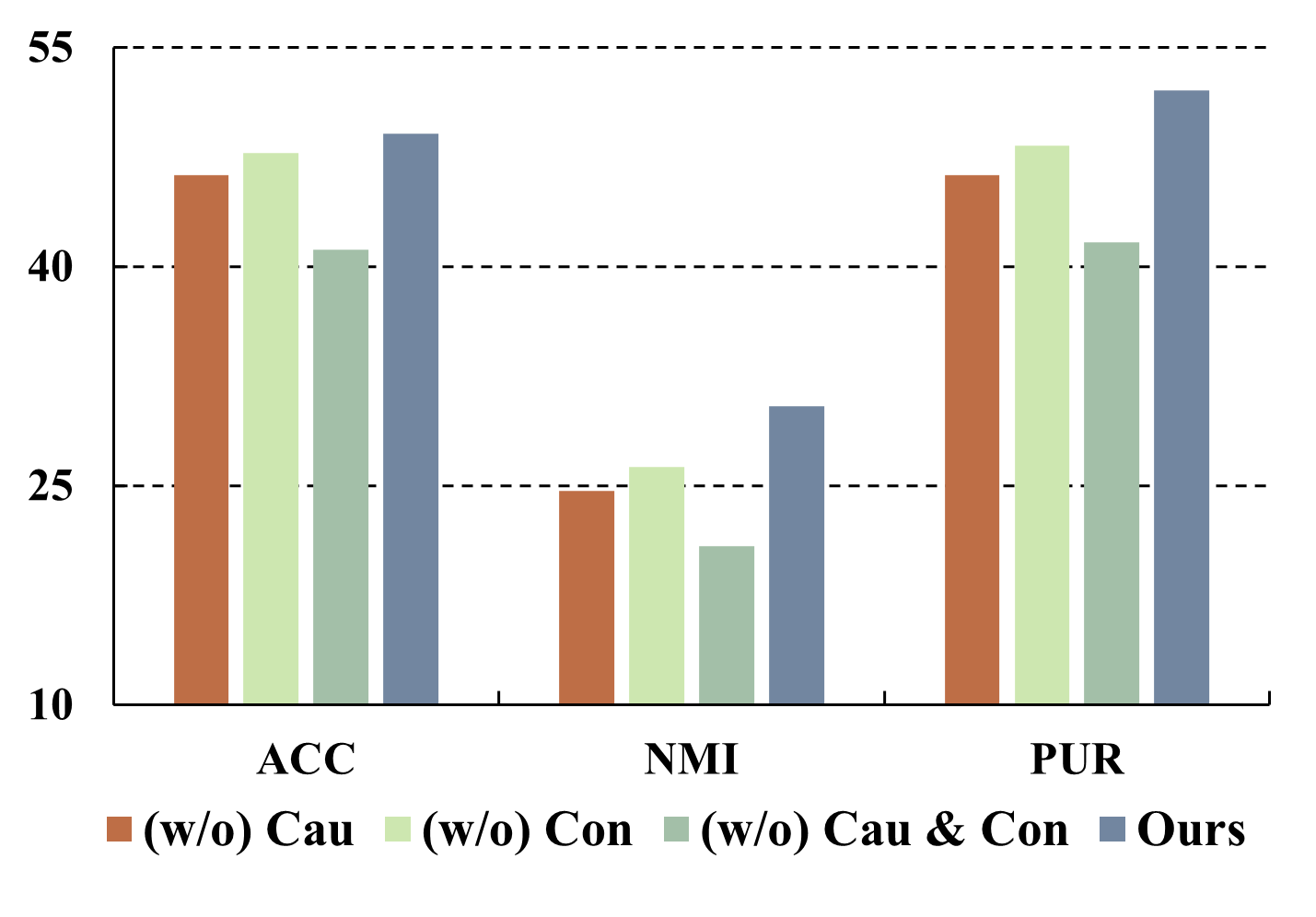}}
\vspace{3pt}
\centerline{{Reuters}}
\end{minipage}
\caption{Ablation studies on eight datasets with fully aligned data. ``(w/o) Cau'', ``(w/o) Con'', ``(w/o) Cau\&Con'', and  ``Ours'' correspond to reduced models by individually removing the causal module, the contrastive regularizer, and all aforementioned modules combined, respectively.}
\label{ablation_all}
\end{figure*}

\begin{figure*}
\centering
\begin{minipage}{0.23\linewidth}
\centerline{\includegraphics[width=1\textwidth]{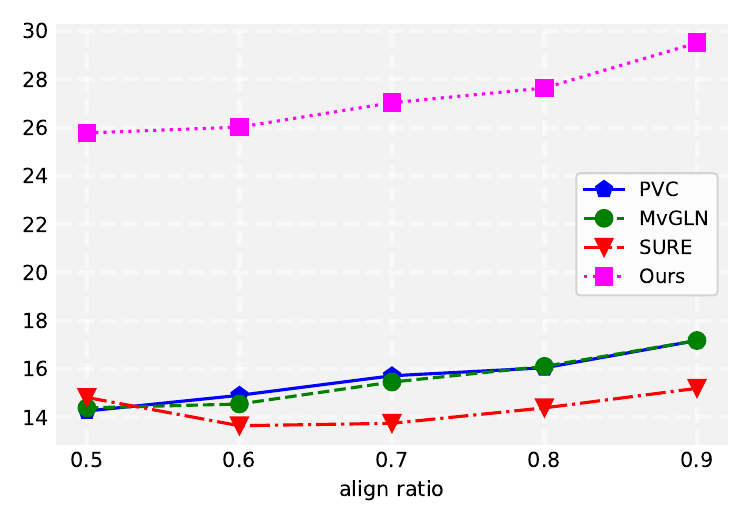}}
\centerline{{Movies}}
\vspace{3pt}
\centerline{\includegraphics[width=1\textwidth]{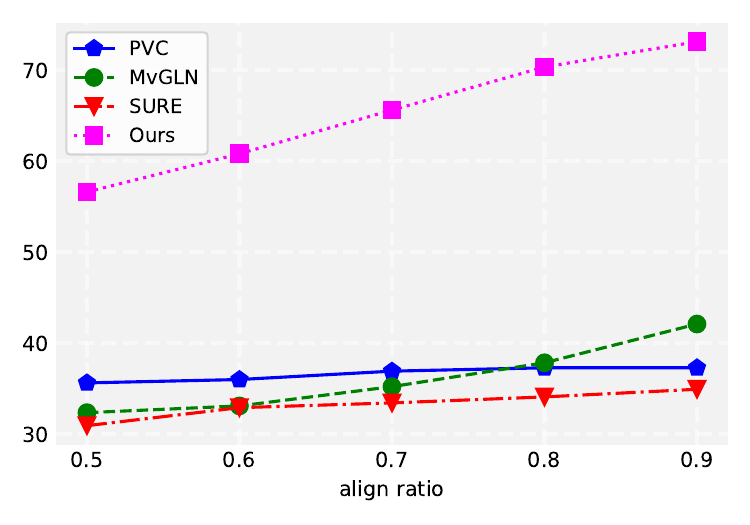}}
\centerline{{BBCSport}}
\vspace{3pt}
\end{minipage}
\begin{minipage}{0.23\linewidth}
\centerline{\includegraphics[width=1\textwidth]{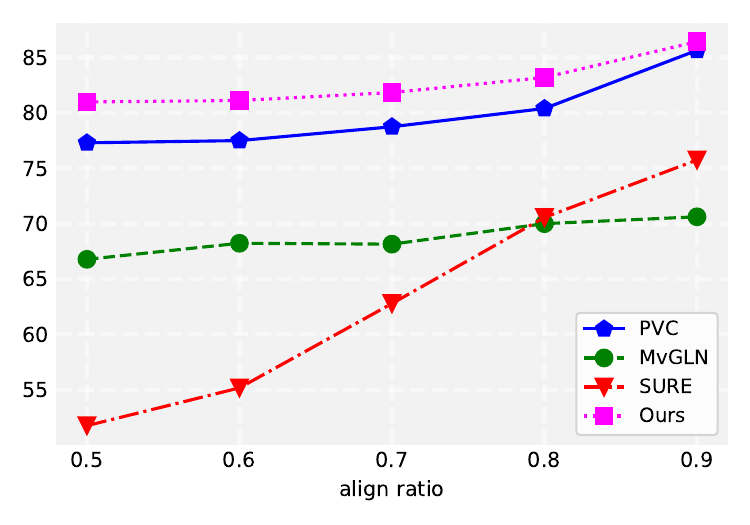}}
\centerline{{UCI-digit}}
\vspace{3pt}
\centerline{\includegraphics[width=1\textwidth]{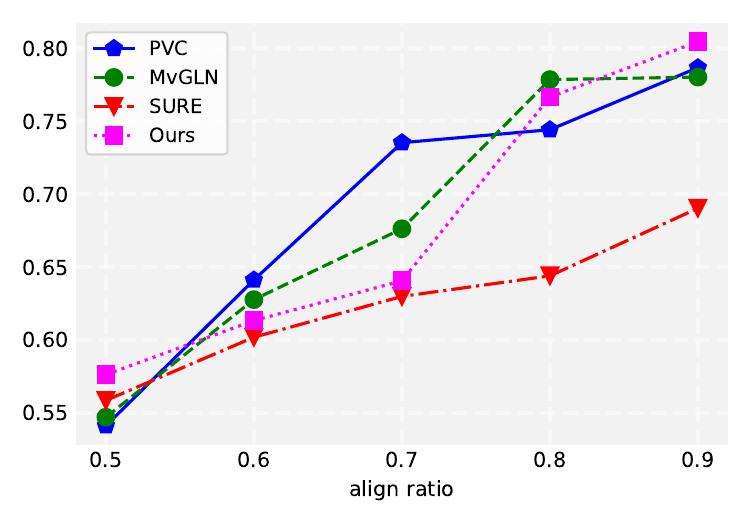}}
\centerline{{WebKB}}
\vspace{3pt}
\end{minipage}
\begin{minipage}{0.23\linewidth}
\centerline{\includegraphics[width=1\textwidth]{Figure/new_align/stl10_acc.pdf}}
\centerline{{STL-10}}
\vspace{3pt}
\centerline{\includegraphics[width=1\textwidth]{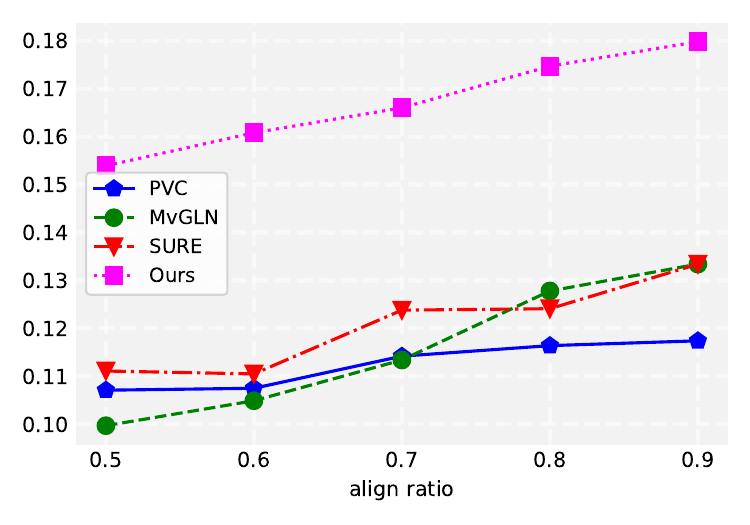}}
\centerline{{SUNRGB-D}}
\vspace{3pt}
\end{minipage}
\begin{minipage}{0.23\linewidth}
\centerline{\includegraphics[width=1\textwidth]{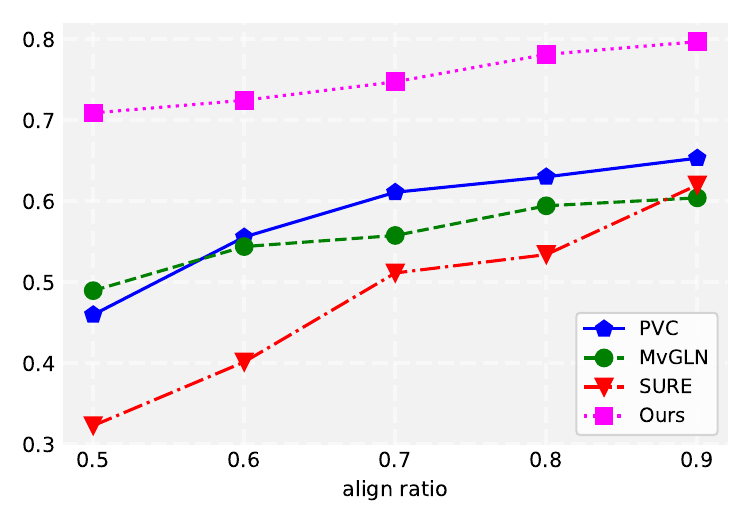}}
\centerline{{Caltech101-7}}
\vspace{3pt}
\centerline{\includegraphics[width=1\textwidth]{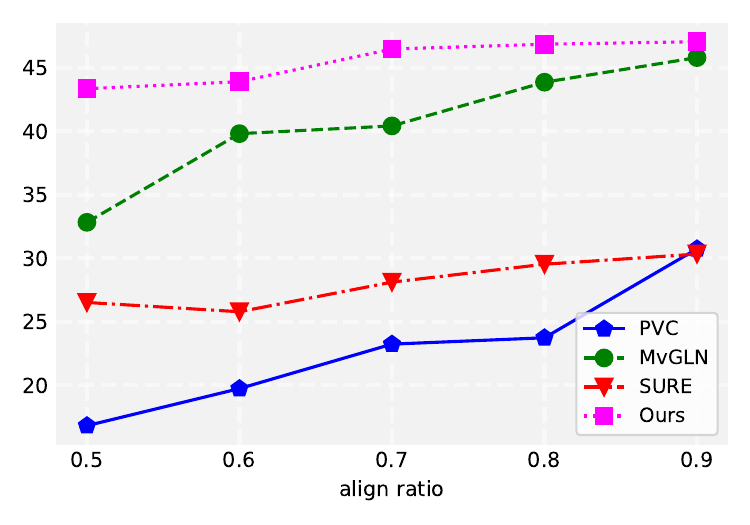}}
\centerline{{Reuters}}
\vspace{3pt}
\end{minipage}
\caption{Clustering performance on eight Datasets with different aligned ratios in accuracy metrics.}
\label{align_acc}
\end{figure*}

\begin{figure*}
\centering
\begin{minipage}{0.23\linewidth}
\centerline{\includegraphics[width=1\textwidth]{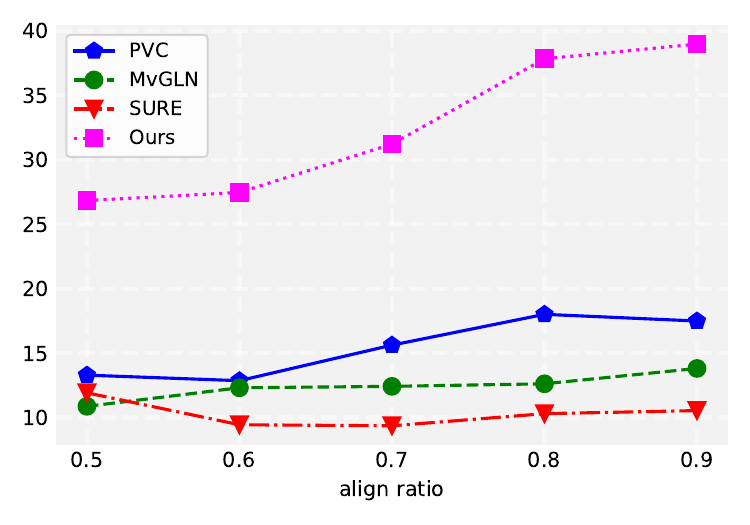}}
\centerline{{Movies}}
\vspace{3pt}
\centerline{\includegraphics[width=1\textwidth]{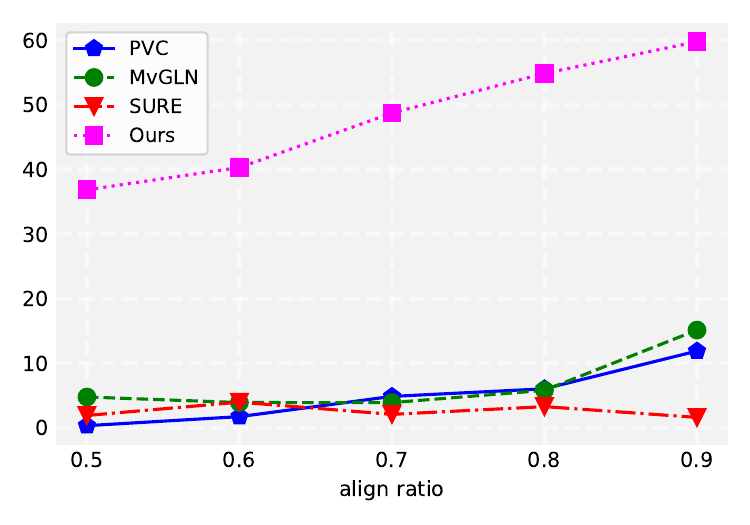}}
\centerline{{BBCSport}}
\vspace{3pt}
\end{minipage}
\begin{minipage}{0.23\linewidth}
\centerline{\includegraphics[width=1\textwidth]{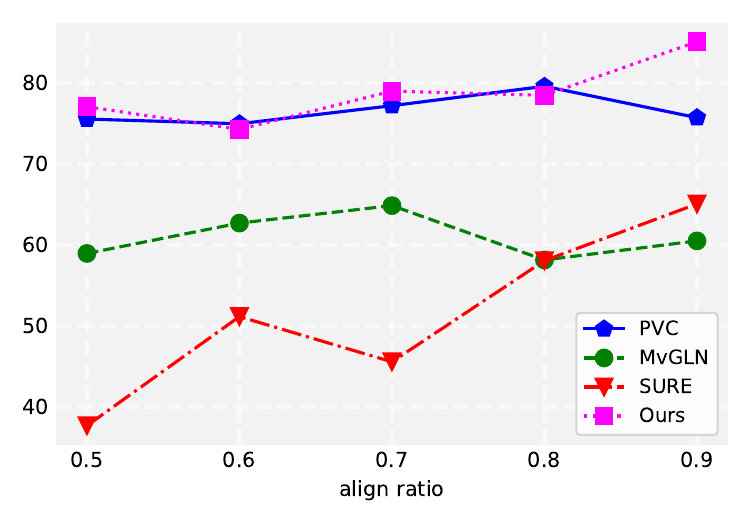}}
\centerline{{UCI-digit}}
\vspace{3pt}
\centerline{\includegraphics[width=1\textwidth]{Figure/new_align/webkb_nmi.pdf}}
\centerline{{WebKB}}
\vspace{3pt}
\end{minipage}
\begin{minipage}{0.23\linewidth}
\centerline{\includegraphics[width=1\textwidth]{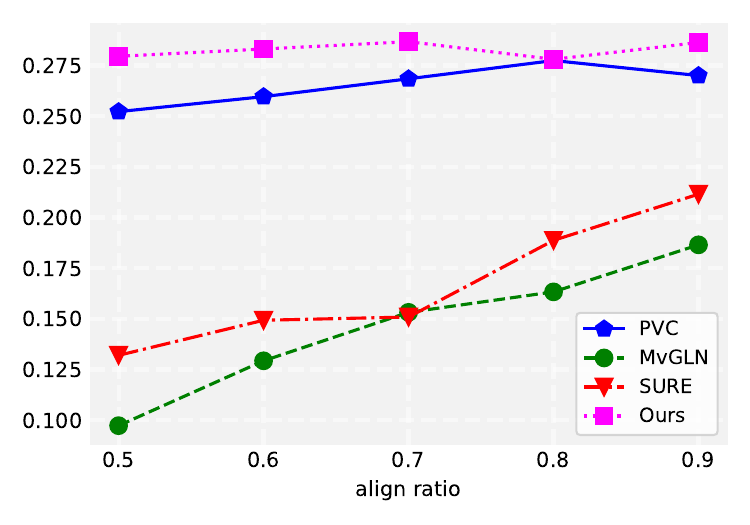}}
\centerline{{STL-10}}
\vspace{3pt}
\centerline{\includegraphics[width=1\textwidth]{Figure/new_align/sund_nmi.pdf}}
\centerline{{SUNRGB-D}}
\vspace{3pt}
\end{minipage}
\begin{minipage}{0.23\linewidth}
\centerline{\includegraphics[width=1\textwidth]{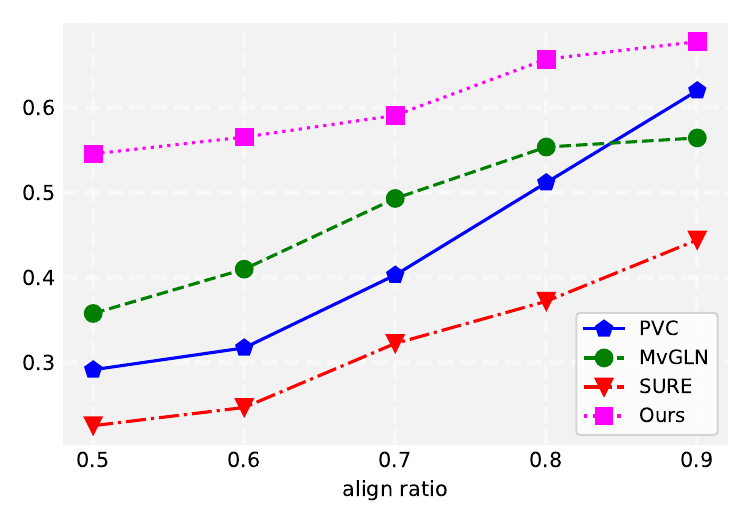}}
\centerline{{Caltech101-7}}
\vspace{3pt}
\centerline{\includegraphics[width=1\textwidth]{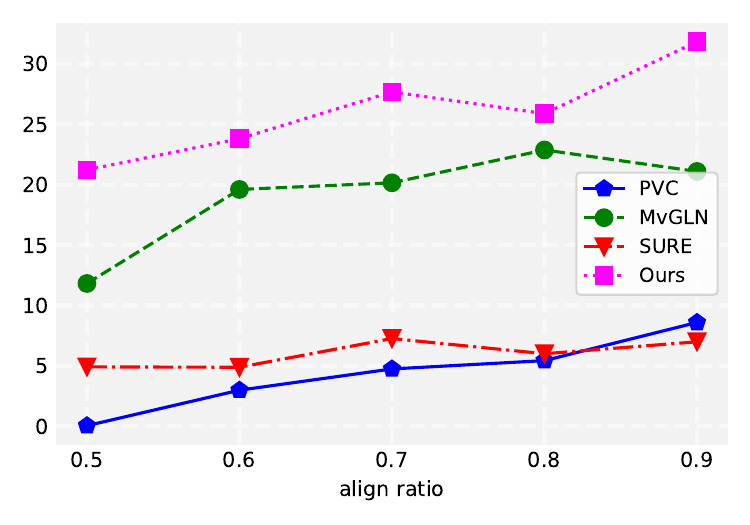}}
\centerline{{Reuters}}
\vspace{3pt}
\end{minipage}
\caption{Clustering performance on eight Datasets with different aligned ratios in NMI metrics.}
\label{align_nmi}
\end{figure*}

\begin{figure*}
\centering
\begin{minipage}{0.23\linewidth}
\centerline{\includegraphics[width=1\textwidth]{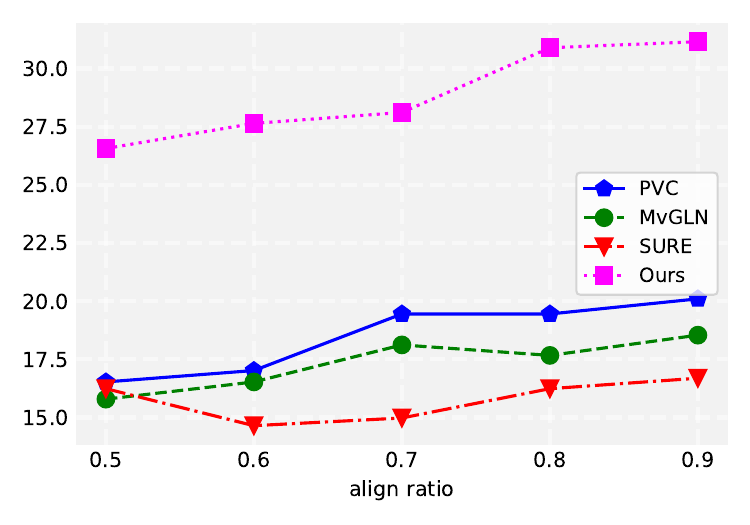}}
\centerline{{Movies}}
\vspace{3pt}
\centerline{\includegraphics[width=1\textwidth]{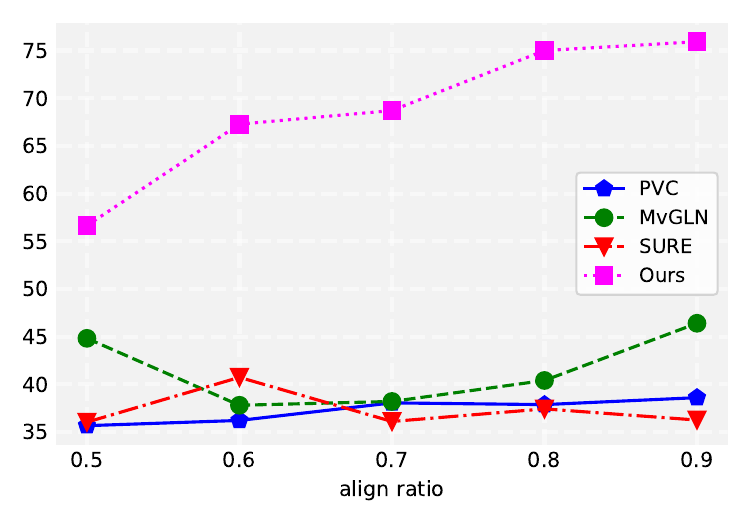}}
\centerline{{BBCSport}}
\vspace{3pt}
\end{minipage}
\begin{minipage}{0.23\linewidth}
\centerline{\includegraphics[width=1\textwidth]{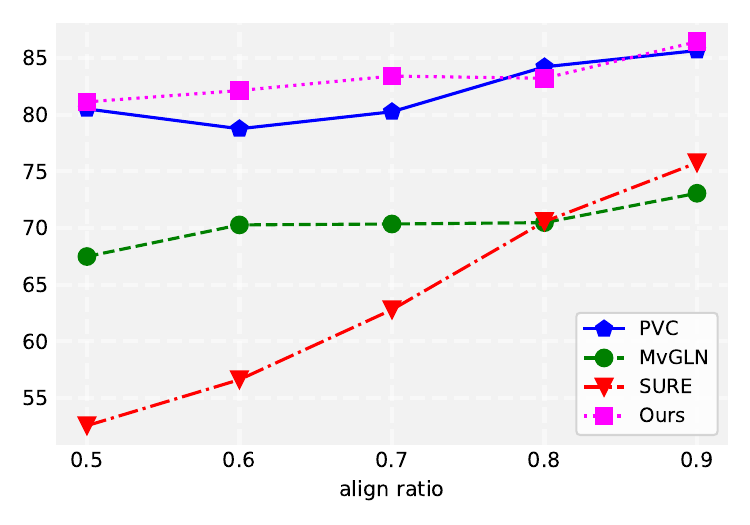}}
\centerline{{UCI-digit}}
\vspace{3pt}
\centerline{\includegraphics[width=1\textwidth]{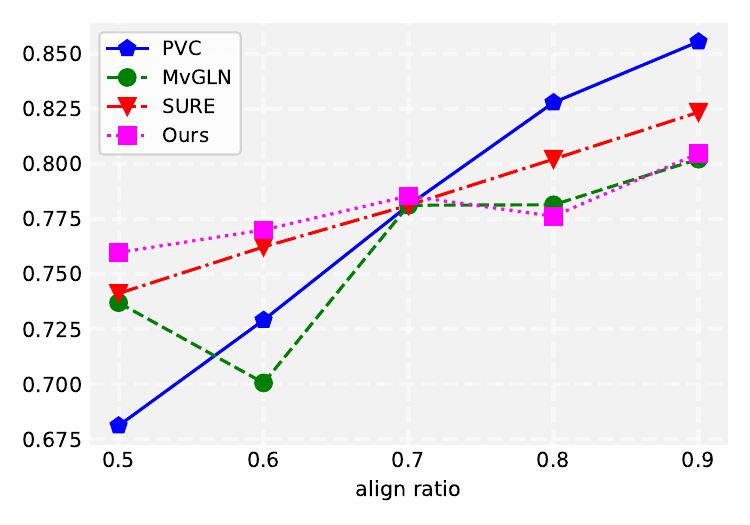}}
\centerline{{WebKB}}
\vspace{3pt}
\end{minipage}
\begin{minipage}{0.23\linewidth}
\centerline{\includegraphics[width=1\textwidth]{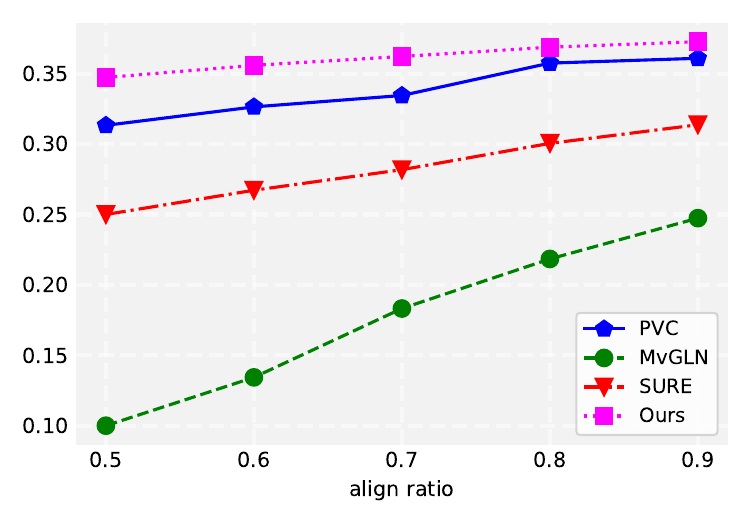}}
\centerline{{STL-10}}
\vspace{3pt}
\centerline{\includegraphics[width=1\textwidth]{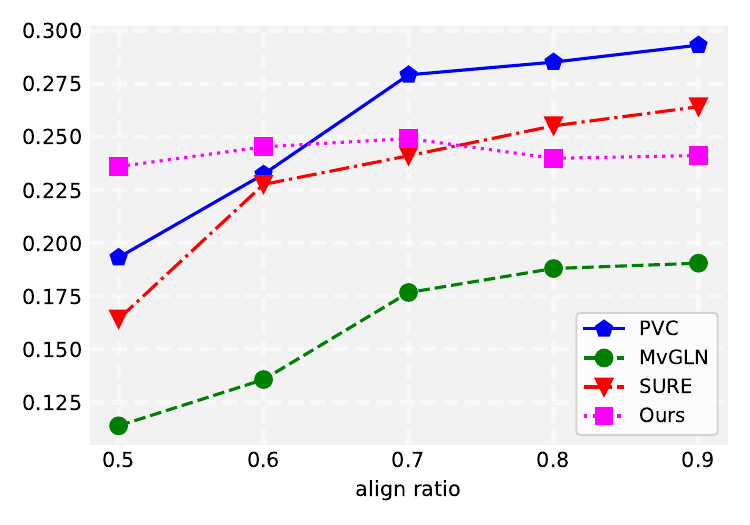}}
\centerline{{SUNRGB-D}}
\vspace{3pt}
\end{minipage}
\begin{minipage}{0.23\linewidth}
\centerline{\includegraphics[width=1\textwidth]{Figure/new_align/cal7_pur.pdf}}
\centerline{{Caltech101-7}}
\vspace{3pt}
\centerline{\includegraphics[width=1\textwidth]{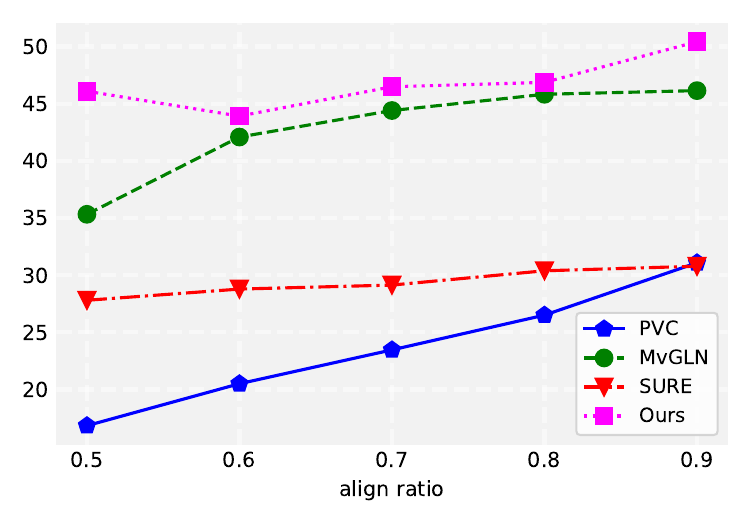}}
\centerline{{Reuters}}
\vspace{3pt}
\end{minipage}
\caption{Clustering performance on eight Datasets with different aligned ratios in PUR metrics.}
\label{align_pur}
\end{figure*}

\begin{figure*}
\centering
\begin{minipage}{0.22\linewidth}
\centerline{\includegraphics[width=1\textwidth]{Figure/tsne/0.png}}
\vspace{5pt}
\centerline{{0 Epoch}}
\centerline{\includegraphics[width=1\textwidth]{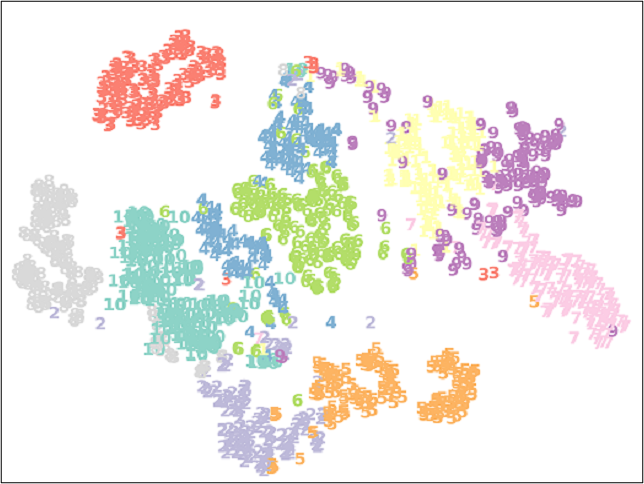}}
\vspace{5pt}
\centerline{{120 Epoch}}
\end{minipage}
\begin{minipage}{0.22\linewidth}
\centerline{\includegraphics[width=1\textwidth]{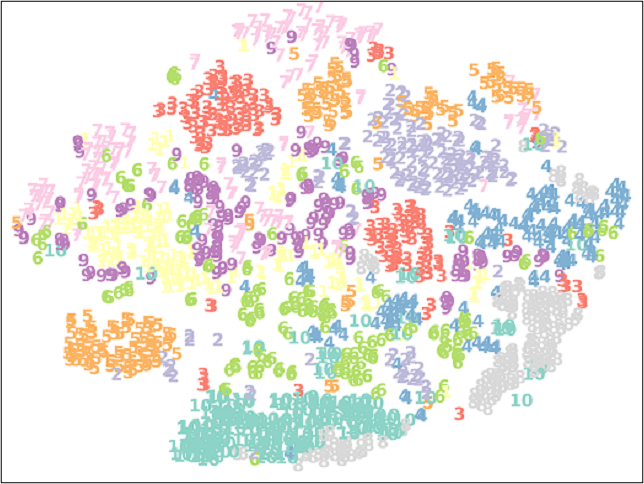}}
\vspace{5pt}
\centerline{{40 Epoch}}
\centerline{\includegraphics[width=1\textwidth]{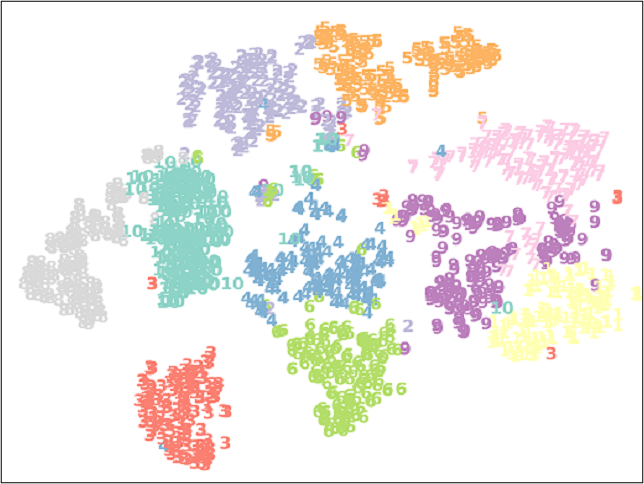}}
\vspace{5pt}
\centerline{{140 Epoch}}
\end{minipage}
\begin{minipage}{0.22\linewidth}
\centerline{\includegraphics[width=1\textwidth]{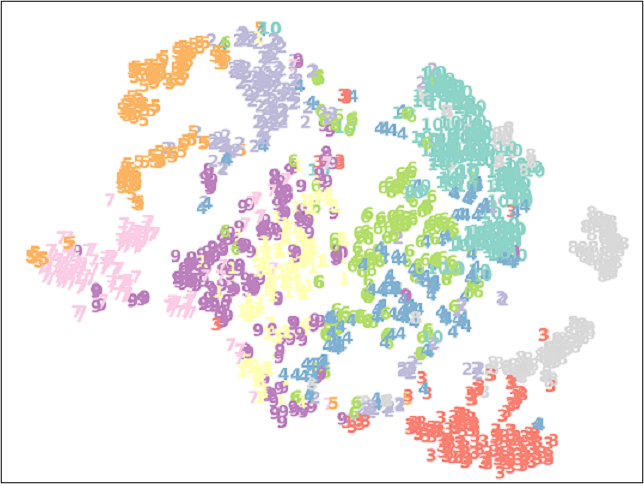}}
\vspace{5pt}
\centerline{{60 Epoch}}
\centerline{\includegraphics[width=1\textwidth]{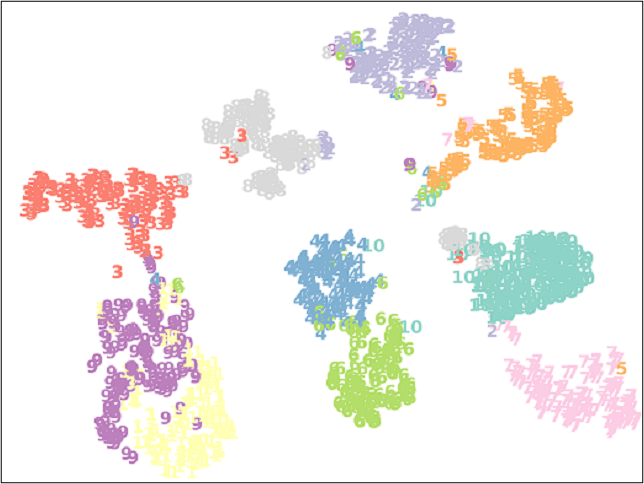}}
\vspace{5pt}
\centerline{{180 Epoch}}
\end{minipage}
\begin{minipage}{0.22\linewidth}
\centerline{\includegraphics[width=1\textwidth]{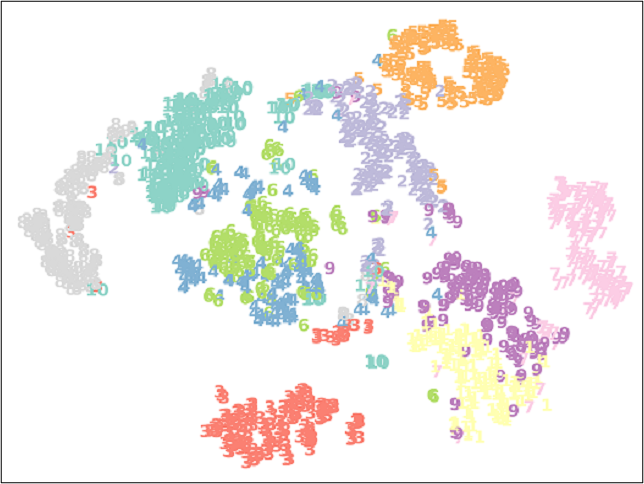}}
\vspace{5pt}
\centerline{{80 Epoch}}
\centerline{\includegraphics[width=1\textwidth]{Figure/tsne/200.png}}
\vspace{5pt}
\centerline{{200 Epoch}}
\end{minipage}
\caption{Visualization of the representations during the training process on UCI-digit dataset.}
\label{t_sne}
\end{figure*}

\section{Additional Experiments}
\subsection{Ablation Studies}

In this section, we first present the ablation study on all datasets under the partially aligned scenario. ``(w/o) Cau'', ``(w/o) Con'', ``(w/o) Cau\&Con'', and  ``Ours'' represent the ablated models where the causal module, the contrastive regularizer, and both modules combined, respectively, are individually removed. In the ``(w/o) Cau\&Con'' configuration, we employ an autoencoder network as the backbone to derive representations for the downstream clustering task. Consistent with the conclusions drawn in the main text, the results are summarized as follows.

\begin{itemize}
  \item We resort the multi-view clustering from the causal perspective. The model generalization is improved when the input is partially aligned data, thus achieving promising performance.

  \item The contrastive module could push the positive sample close, and pull the negative sample away, enhancing the model's discriminative capacity. The model could achieve better clustering outcomes.
  
\end{itemize}

Besides, we perform ablation studies with fully aligned data to assess the effectiveness of our designed modules, namely the causal module and contrastive regularizer. Specifically, "(w/o) Cau," "(w/o) Con," "(w/o) Cau\&Con," and "Ours" denote reduced models created by individually omitting the causal module, the contrastive regularizer, and both modules together. In this paper, we employ an autoencoder network as the core architecture to derive representations for the subsequent clustering task, referred to as "(w/o) Cau\&Con,". The outcomes are depicted in Fig.~\ref{ablation_all}. These results clearly demonstrate that the exclusion of any of the designed modules leads to a significant decrease in clustering performance, underscoring the essential role each module plays in optimizing overall performance.

\subsection{Different Align Ratio}

To evaluate the performance of CauMVC under different alignment ratios, we conduct experiments on eight datasets, with the results presented in Fig.~\ref{align_acc}, Fig.~\ref{align_nmi}, and Fig.~\ref{align_pur}. The results clearly demonstrate that CauMVC outperforms other baseline models across various alignment ratios in most scenarios. This highlights its strong generalization capability in handling partially aligned data effectively.

\subsection{Sensitivity Analysis of $\alpha$ and $\beta$}

To further investigate the impact of the parameters \( \alpha \) and \( \beta \) on our model, we conduct experiments on the BBCSport dataset, analyzing parameter values within the range of \(\{0.01, 0.1, 1.0, 10, 100\}\). Due to space constraints, the experimental results for the BBCSport dataset are provided in the Appendix. Based on the results presented in Fig.~\ref{sen_alpha}, we draw the following observations:  

\begin{itemize}
    \item When \( \alpha \) and \( \beta \) are assigned extreme values (0.1 or 100), the clustering performance tends to degrade. We hypothesize that this decline results from an imbalance in the loss function. Moreover, the model achieves optimal performance when the trade-off parameters are set around 1.0. 

    \item The results also indicate that \( \alpha \) has a more significant impact on model performance, suggesting that the causal model plays a crucial role in enhancing the overall effectiveness of the approach.
\end{itemize}

{
    \small
    \bibliographystyle{ieeenat_fullname}
    \bibliography{main}
}

\end{document}